\documentclass{amsart}
\usepackage[utf8]{inputenc}
\usepackage{amsmath}
\usepackage{graphicx}
\usepackage{bbm}
\usepackage{hyperref}
\usepackage{caption}
\usepackage{tipa}
\usepackage{xspace}
\usepackage[frozencache, cachedir=minted-cache]{minted}

\AtBeginDocument{%
  }

\setminted[python]{bgcolor=bg,fontsize=\footnotesize,autogobble=True,python3=true}

\newtheorem{proposition}{Proposition}[subsubsection]
\theoremstyle{definition}
\newtheorem{definition}{Definition}[subsubsection]

\newcommand*{\codeobj}[1]{\texttt{#1}\xspace}
\newcommand{\R}{\mathbb R}
\newcommand{\E}{\mathbb E}

\begin{document}
\definecolor{bg}{rgb}{0.95,0.95,0.95}

\title{Parametric information geometry with the package Geomstats}
\author{Alice Le Brigant, Jules Deschamps, Antoine Collas and Nina Miolane}
\date{}

\maketitle

\begin{abstract}
    We introduce the information geometry module of the Python package Geomstats. The module first implements Fisher-Rao Riemannian manifolds of widely used parametric families of probability distributions, such as normal, gamma, beta, Dirichlet distributions, and more. The module further gives the Fisher-Rao Riemannian geometry of any parametric family of distributions of interest, given a parameterized probability density function as input. The implemented Riemannian geometry tools allow users to compare, average, interpolate between distributions inside a given family. Importantly, such capabilities open the door to statistics and machine learning on probability
    distributions. We present the object-oriented implementation of the module along with illustrative examples and show how it can be used to perform learning on manifolds of parametric probability distributions.
\end{abstract}

\section{Introduction}

Geomstats \cite{miolane2020geomstats} is an open-source Python package for statistics and learning on manifolds. Geomstats allows users to analyze complex data that belong to manifolds equipped with various geometric structures, such as Riemannian metrics. This type of data arise in many applications: in computer vision, the manifold of 3D rotations models movements of articulated objects like the human spine or robotics arms \cite{Arsigny:PHD:2006}; and in biomedical imaging, biological shapes are studied as elements of shape manifolds \cite{Dryden1998,Younes2012Spaces}. The manifolds implemented in Geomstats come equipped with Riemannian metrics that allow users to compute distances and geodesics, among others. Geomstats also provides statistical learning algorithms that are compatible with the Riemannian structures, \textit{i.e.}, that can be used in combination with any of the implemented Riemannian manifolds. These algorithms are geometric generalizations of common estimation, clustering, dimension reduction, classification and regression methods to nonlinear manifolds.

Probability distributions are a type of complex data often encountered in applications: in text classification, multinomial distributions are used to represent documents by indicating words frequencies \cite{lebanon2012learning}; in medical imaging, multivariate normal distributions are used to model diffusion tensor images \cite{lenglet2006}. Many more examples of applications can be found in the rest of this paper. Spaces of probability distributions possess a nonlinear structure that can be captured by two main geometric representations: one provided by optimal transport and one arising from information geometry \cite{amari2016information}. In optimal transport, probability distributions are seen as elements of an infinite-dimensional manifold equipped with the Otto-Wasserstein metric \cite{otto2001geometry,ambrosio2013user}. By contrast, information geometry gives a finite-dimensional manifold representation of parametric families of distributions. \footnote{There also exists a non parametric-version that can be defined on the infinite-dimensional space of probability distributions \cite{friedrich1991fisher}, that we do not consider here.} 

Specifically, information geometry represents the probability distributions of a given parametric family by their parameter space, on which the Fisher information is used to define a Riemannian metric \textemdash the so-called \textit{Fisher-Rao metric} or \textit{Fisher information metric} \cite{rao1945}. This metric is a powerful tool to compare and analyze probability distributions inside a given parametric family. It is invariant to diffeomorphic changes of parametrization, and it is the only metric invariant with respect to sufficient statistics, as proved by Cencov \cite{cencov1982}. Most importantly, the Fisher-Rao metric comes with Riemannian geometric tools such as geodesics, geodesic distance and intrinsic means, that give an intrinsic way to interpolate, compare, average probability distributions inside a given parametric family. By construction, geodesics and means for the Fisher-Rao metric never leave the parametric family of distributions, contrary to their Wasserstein-metric counterparts. These intrinsic computations can then serve as building blocks to apply learning algorithms to parametric probability distributions.

The geometries of several parametric families have been studied in the literature, and some relate to well-known Riemannian structures: the Fisher-Rao geometry of univariate normal distributions is hyperbolic \cite{atkinson1981}; the Fisher-Rao geometry of multinomial distributions is spherical \cite{kass1989geometry}; and the Fisher-Rao geometry of multivariante distributions of fixed mean coincides with the affine-invariant metric on the space of symmetric positive definite matrices \cite{pennec2006riemannian}. 

\subsubsection*{Contributions.}
Computational tools for optimal transport have been proposed, in Python in particular \cite{flamary2021pot}. However, to the best of our knowledge, there exists no wide-ranging open source Python implementation of parametric information geometry, despite a recent implementation in Julia \cite{arutjunjan2021}. To fill this gap, this paper presents a module of Geomstats that implements the Fisher-Rao geometries of standard parametric families of probability distributions. Each parametric family of distributions is implemented through its Fisher-Rao manifold with associated exponential and logarithm maps, geodesic distance and geodesics. These manifolds are compatible with the statistical learning algorithms of Geomstats' learning module, which can therefore be applied to probability distributions data. As in the rest of Geomstats, the implementation is object-oriented and extensively unit-tested. All operations are vectorized for batch computation and support is provided for different execution backends — namely NumPy, Autograd, PyTorch, and TensorFlow.

\subsubsection*{Outline.}
The rest of the paper is organized as follows. Section~\ref{sec:module} provides the necessary background of Riemannian geometry and introduces the structure of Geomstats' information geometry module, \textit{i.e.}, the Python classes used to define a Fisher-Rao geometry. Section~\ref{sec:catalogue} details the geometries of the parametric families implemented in the module, along with code illustrations and examples of real-world usecases in the literature. Section~\ref{sec:application} presents an application of the information geometry tools of Geomstats to geometric learning on probability distributions. Altogether, the proposed information geometry module represents the first comprehensive implementation of parametric information geometry in Python.

\section{Information geometry module of geomstats}\label{sec:module}

\begin{figure}
    \centering
    \centerline{
    \includegraphics[scale=.45]{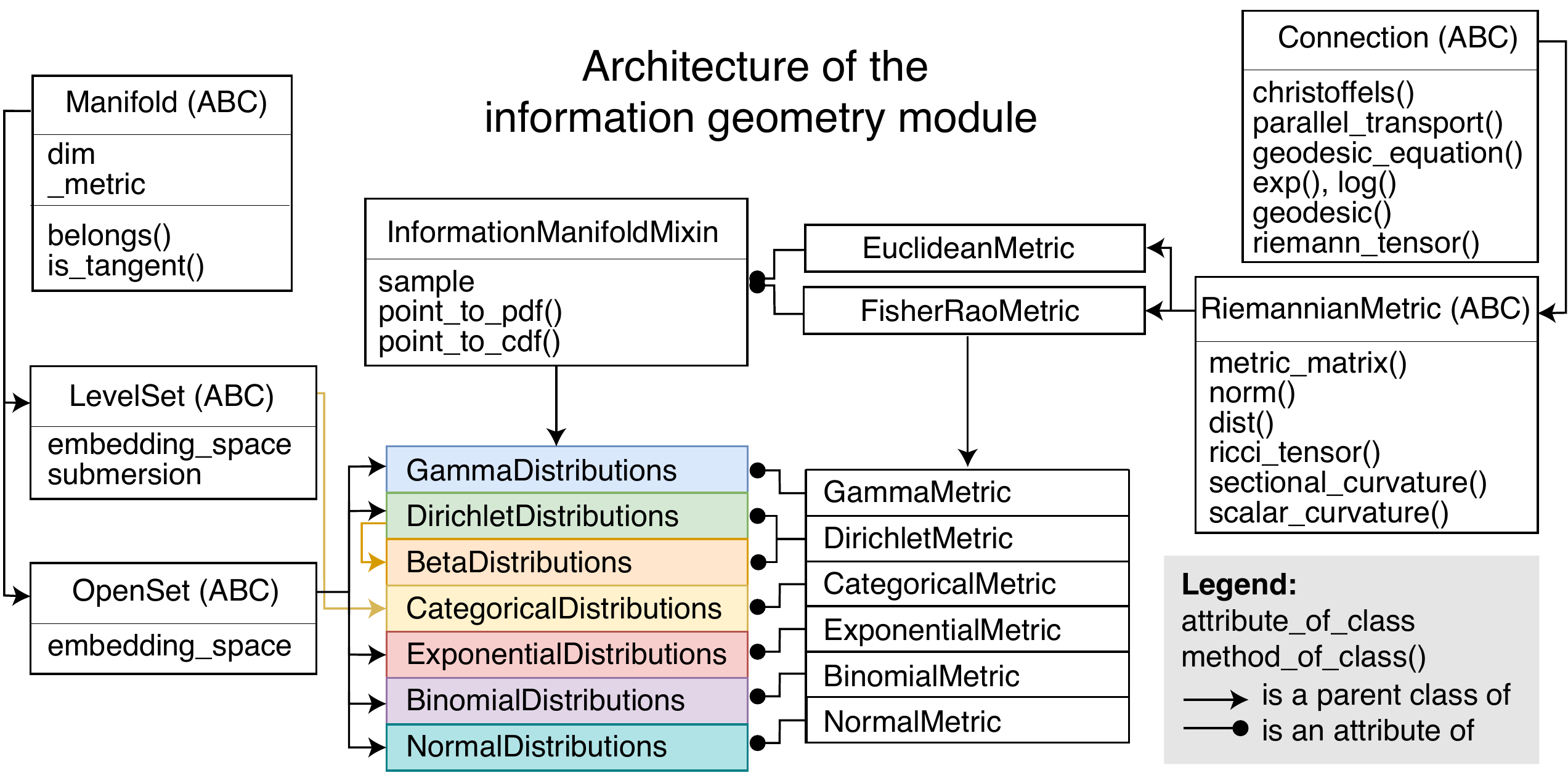}}
    \smallskip
    \caption{Architecture of the information geometry module of \codeobj{Geomstats}. The \codeobj{InformationManifold} Python mixin and the \codeobj{FisherRaoMetric} Python class implement the building blocks of parametric information geometry. The most common parametric families of distributions are Python classes represented in colors, and inherit from the \codeobj{InformationManifold} mixin. They are equipped with their respective Riemannian metrics, which themselves inherit from the \codeobj{FisherRaoMetric} class. The abstract (ABC) Python classes \codeobj{Manifold}, \codeobj{OpenSet}, \codeobj{LevelSet}, \codeobj{Connection}, \codeobj{RiemannianMetric} provides tools of Riemannian geometry to compute on the information manifolds.}
    \label{fig:architecture}
\end{figure}

This section describes the design of the \codeobj{information\_geometry} module and its integration into Geomstats.
The proposed module implements a Riemannian manifold structure for common parametric families of probability distributions, such as normal distributions, using the object-oriented architecture shown in Fig.~\ref{fig:architecture}. The Riemannian manifold structure is encoded by two Python classes: one for the parameter manifold of the family of distributions and one for the Fisher-Rao metric on this manifold. For example, in the case of normal distributions, these Python classes are called \codeobj{NormalDistributions} and \codeobj{NormalMetric}. They inherit from more general Python classes, in particular the \codeobj{Manifold}, \codeobj{Connection} and \codeobj{RiemannianMetric} classes. These are abstract classes that define structure, but cannot be instantiated, contrary to their child classes \codeobj{NormalDistributions} and \codeobj{NormalMetric}. They also inherit from the \codeobj{InformationManifold} mixin and the \codeobj{FisherRaoMetric}: these are Python structures specific to the information geometry module. This section details this architecture along with some theoretical background. For more details on Riemannian geometry, we refer the interested reader to a standard textbook such as \cite{do1992riemannian}.

\subsection{Manifold}

The \codeobj{Manifold} abstract class implements the structure of a \textit{manifold}, \textit{i.e.}, a space that locally resembles a vector space, without necessarily having its global flat structure. Manifolds of dimension $d$ can be defined in an abstract way, \textit{i.e.}, without considering their embedding in an ambient space, by ``gluing'' together small pieces of Euclidean space $\R^d$ using charts. We will only consider smooth manifolds, for which the transition from one chart to another is smooth. In addition, submanifolds of a larger Euclidean space $\R^N$ can be defined locally in various ways: \textit{e.g.}, using a parametrization, an implicit function or as the graph of a function \cite{guigui2023}. The simplest examples of manifolds are Euclidean spaces $\R^d$, or more generally vector spaces in finite dimensions, open sets of vector spaces (there is only one chart which is the identity) and level sets of functions (defined globally by one implicit function). These important cases are implemented in the abstract classes \codeobj{VectorSpace}, \codeobj{OpenSet} and \codeobj{LevelSet}, which are child classes of \codeobj{Manifold} as shown in Figure~\ref{fig:architecture}.

A $d$-dimensional manifold $M$ admits a \textit{tangent space} $T_xM$ at each point $x\in M$ that is a $d$-dimensional vector space. For open sets of $\R^d$, it can be identified with $\R^d$ itself. 
The classes that inherit from \codeobj{Manifold} contain methods that allow users to verify that an input is a point belonging to the manifold via the \codeobj{belongs()} method or that an input is a tangent vector to the manifold at a given base point via the method \codeobj{is\_tangent()} (see Figure~\ref{fig:architecture}).

\subsection{Connection}

The \codeobj{Connection} class implements the structure of an affine connection, which is a geometric tool that defines the generalization of straight lines, addition and subtraction to nonlinear manifolds. To this end, a connection allows us to take derivatives of vector fields, \textit{i.e.}, mappings $V:M\rightarrow TM$ that associate to each point $p$ a tangent vector $V(p)\in T_pM$. Precisely, an \textit{affine connection} is a functional $\nabla$ that acts on pairs of vector fields $(U,V) \mapsto \nabla_UV$ according to the following rules: for any vector fields $U,V,W$ and differentiable function $f$,
\begin{align*}
\nabla_{fU+V} W &= f\nabla_UV + \nabla_UW,\\
\nabla_U(fV+W) &= U(f)V + f\nabla_UV + \nabla_UW,
\end{align*}
where $U(f)$ denotes the action of the vector field $U$ on the differentiable function $f$. The action induced by the connection $\nabla$ is referred to as \textit{covariant derivative}.  

\subsubsection*{Geodesics}

If $\gamma(t)$ is a curve on $M$, its velocity $\dot\gamma(t)$ is a vector field along $\gamma$, i.e. $\dot\gamma(t)\in T_{\gamma(t)}M$ for all $t$. The acceleration of a curve is therefore the covariant derivative of this velocity field with respect to the affine connection $\nabla$. A curve $\gamma$ of zero acceleration 
\begin{equation}
    \label{general_geod_eq}
    \nabla_{\dot\gamma}\dot\gamma=0,
\end{equation}
is called a \textit{$\nabla$-geodesic}. Geodesics are the manifolds counterparts of vector spaces' straight lines. Equation~\eqref{general_geod_eq} translates into a system of ordinary differential equations (ODEs) for the coordinates of the geodesic $\gamma=(\gamma_1,\hdots,\gamma_d)$
\begin{equation}\label{geod_eq}
    \ddot\gamma_k+\sum_{i,j=1}^d\Gamma_{ij}^k(\gamma)\dot\gamma_i\dot\gamma_j=0, \quad k=1,\hdots,d,
\end{equation}
where the coefficients $\Gamma_{ij}^k$ are the \textit{Christoffel symbols} that define the affine connection in local coordinates. In the \codeobj{Connection} class, Equation~\eqref{geod_eq} is implemented in the \codeobj{geodesic\_equation()} method and the Christoffel symbols are implemented in the \codeobj{christoffels()} method (see Figure~\ref{fig:architecture}).  

\subsubsection*{Exp and Log maps}
Existence results for solutions of ODEs allow us to define geodesics starting at a point $x$ with velocity $v\in T_xM$ for times $t$ in a neighborhood of zero, or equivalently for all time $t\in[0,1]$ but for tangent vectors $v$ of small norm. The \textit{exponential map} at $x\in M$ associates to any $v\in T_xM$ of sufficiently small norm the end point $\gamma(1)$ of a geodesic $\gamma$ starting from $\theta$ with velocity $v$:
$$\exp_x(v)=\gamma(1), \quad \text{where} \begin{cases} 
\gamma \text{ is a geodesic}, \\ 
\gamma(0) = x, \, \dot\gamma(0) = v.
\end{cases}$$
If $B$ is a small ball of the tangent space $T_xM$ centered at $0$ on which $\exp_x$ is defined, then $\exp_x$ is a diffeomorphism from $B$ onto its image and its inverse $\log_x \equiv \exp_{x}^{-1}$ defines the \textit{logarithm map}, which associates to any point $y$ the velocity $v\in T_xM$ necessary to get to $y$ when departing from $x$:
$$\log_{x}(y)=v \quad\text{where}\quad  \exp_x(v)=y.$$
The exponential and logarithm maps can be seen as generalizations of the Euclidean addition and subtraction to nonlinear manifolds. Both maps are implemented in the \codeobj{exp()} and \codeobj{log()} methods of the \codeobj{Connection} class, which further allow us to get other tools
such as \codeobj{parallel\_transport()} (see Figure~\ref{fig:architecture}). We refer to \cite{guigui2023} for additional details on the \codeobj{Connection} class.

\subsection{Riemannian metric}

Just like there is an abstract Python class that encodes the structure of manifolds, the abstract class \codeobj{RiemannianMetric} encodes the structure of Riemannian metrics. A \textit{Riemannian metric} is a collection of inner products $(\langle\cdot,\cdot\rangle_p)_{p\in M}$ defined on the tangent spaces of a manifold $M$, that depend on the base point $p\in M$ and varies smoothly with respect to it. 

\subsubsection*{Levi-Civita Connection} A Riemannian metric is associated with a unique affine connection, called the \textit{Levi-Civita connection}, which is the only affine connection that is symmetric and compatible with the metric, \textit{i.e.}, that verifies
\begin{align*}
    UV-VU = \nabla_UV - \nabla_VU\\
    U\langle V,W\rangle = \langle \nabla_UV, W\rangle + \langle V,\nabla_UW\rangle
\end{align*}
for all vector fields $U,V,W$. The geodesics of a Riemannian manifold are those of its Levi-Civita connection. The class \codeobj{RiemannianMetric} is therefore a child class of \codeobj{Connection} and inherits all its methods, including \codeobj{geodesic()}, \codeobj{exp()} and \codeobj{log()}. The class \codeobj{RiemannianMetric} overwrites the \codeobj{Connection} class' method \codeobj{christoffels()} and computes the Christoffel symbols using derivatives of the metric. The geodesics, by the compatibility property, have velocity of constant norm, \textit{i.e.}, are parametrized by arc length.

\subsubsection*{Geodesic Distance}
The \codeobj{dist()} method implements the geodesic distance induced by the Riemannian metric, defined between two points $x,y\in M$ to be the length of the shortest curve linking them, where the length of a (piecewise) smooth curve $\gamma:(0,1)\rightarrow M$ is computed by integrating the norm of its velocity
$$d(x,y)=\inf_{\gamma;\gamma(0)=x, \gamma(1)=y}L(\gamma),\quad \text{where}\quad L(\gamma)=\int_0^1 || \dot\gamma(t) ||_{\gamma(t)} dt,$$
using the norm induced by the Riemannian metric. 
In a Riemannian manifold, geodesics extend another property of straight lines: they are locally length-minimizing. In a geodesically complete manifold, any pair of points can be linked by a minimizing geodesic, not necessarily unique, and the \codeobj{dist()} can be computed using the $\codeobj{log}$ map: $$\forall x,y \in M,\quad d(x, y) = ||\log_x(y)||_x.$$

\subsubsection*{Curvatures}
Finally, different notions of curvature are implemented, including the \codeobj{riemann\_curvature()} tensor and \codeobj{sectional\_curvature()}, among others (see Figure~\ref{fig:architecture}). The Riemann curvature tensor is defined from the connection, namely for any vector fields $U,V,W$ as $R(U,V)W = \nabla_{[U,V]}W+\nabla_V\nabla_UW-\nabla_U\nabla_VW$. Sectional curvature at $x\in M$ is a generalization of the Gauss curvature of a surface in $\R^3$. It is defined for any two-dimensional subspace $\sigma(u,v) \subset T_xM$ spanned by tangent vectors $u,v$, as
$$K_{\sigma(u,v)}(x)=\frac{\langle R(u, v)v,u\rangle}{\langle u,u\rangle \langle v,v\rangle-\langle u,v\rangle^2}.$$
It yields important information on the behavior of geodesics, since a geodesically complete and simply connected manifold with everywhere negative sectional curvature (a \textit{Hadamard manifold}) is globally diffeomorphic to $\R^d$ through the exponential map. Consequently, negatively curved spaces share some of the nice properties of Euclidean spaces: any two points can be joined by a unique minimizing geodesic, the length of which gives the geodesic distance.


\subsection{Information manifold}

The proposed \codeobj{information\_geometry} module is integrated into the differential geometry structures implemented in Geomstats. The module contains child classes of \codeobj{Manifold} that represent parametric families of probability distributions, and child classes of \codeobj{RiemannianMetric} that define the Fisher information metric on these manifolds. The combination of two such classes define what we call an \textit{information manifold,} which is specified by an inheritance from the mixin: \codeobj{InformationManifoldMixin} shown in Figure~\ref{fig:architecture}.

\subsubsection*{Parameter Manifolds}
Specifically, consider a family of probability distributions on a space $\mathcal X$, typically $\mathcal X=\R^n$ for some integer $n$. Assume that the distributions in the family are absolutely continuous with respect to a reference measure $\lambda$ (such as the Lebesgue measure on $\R^n$) with densities 
$$f(x|\theta), \quad x\in\mathcal X, \theta\in\Theta,$$
with respect to $\lambda$, where $\theta$ is a parameter belonging to $\Theta$ an open subset of $\R^d$. Then, this parametric family is represented by the \textit{parameter manifold $\Theta$}. The \codeobj{information\_geometry} module implements this manifold as a child class of one of the abstract classes \codeobj{OpenSet} and \codeobj{LevelSet}, which are themselves children of \codeobj{Manifold}. Most of the parameter manifolds are implemented as child classes of \codeobj{OpenSet} as shown in Figure~\ref{fig:architecture}. Other parameter manifolds are implemented more easily with another class. This is the case of \codeobj{CategoricalDistributions}, which inherits from \codeobj{LevelSet} as its parameter space is the interior of the simplex.

\subsubsection*{Information Manifolds}
Parameter manifolds also inherit from the mixin class, called \codeobj{InformationManifoldMixin}, which turns them into \textit{information manifolds}. First, this mixin endows them with specific methods such as \codeobj{sample()}, which returns a sample of the distribution associated to a given parameter $\theta \in \Theta$, or \codeobj{point\_to\_pdf()}, which returns the probability density function (or probability mass function) associated to a given parameter $\theta \in \Theta$ (see Figure~\ref{fig:architecture}). 

For example, to generate at random a categorical distribution on a space of $5$ outcomes, we instantiate an object of the class \codeobj{CategoricalDistributions} with dimension $4$ using \codeobj{manifold = CategoricalDistributions(4)} and define \codeobj{parameter = manifold.random\_point()}. Then, in order to sample from this distribution, one uses \codeobj{samples = manifold.sample(parameter, n\_samples=10)}.


Second, the \codeobj{InformationManifoldMixin} endows the parameter manifolds with a Riemannian metric defined using the Fisher information, called the \textit{Fisher-Rao metric} and implemented in the \codeobj{FisherRaoMetric} class shown in Figure~\ref{fig:architecture}. The Fisher information is a notion from statistical inference that measures the quantity of information on the parameter $\theta$ contained in an observation with density $f(\cdot,\theta)$. It is defined, under certain regularity conditions \cite{lehmann2006theory}, as
\begin{equation}
    \label{eq:fisherinfo}
    I(\theta)=-\E_\theta\left[\mathrm{Hess}_\theta\left(\log f(X|\theta)\right)\right],
\end{equation}
where $\mathrm{Hess}_\theta$ denotes the hessian with respect to $\theta$ and $\E_\theta$ is the expectation taken with respect to the random variable $X$ with density $f(\cdot, \theta)$. If this $d$-by-$d$ matrix is everywhere definite, it provides a Riemannian metric on $\Theta$, called the Fisher-Rao metric, where the inner product between two tangent vectors $u,v$ at $\theta\in\Theta$ is defined by
\begin{equation}
    \label{eq:FRmetric}
    \langle u,v\rangle_{\theta}=u^\top I(\theta) v.
\end{equation}
Here the tangent vectors $u, v$ are simply vectors of $\R^d$ since $\Theta$ is an open subset of $\R^d$. In the sequel, we will describe the Fisher-Rao metric for different parametric statistical families by providing the expression of the infinitesimal length element 
$$ds^2= \langle d\theta, d\theta\rangle_\theta = d\theta^\top I(\theta)d\theta$$
The metric matrix $I$ is implemented using automatic differentiation in the \codeobj{FisherRaoMetric} class. This allows users to get the Fisher-Rao Metric of any parametric family of probability distributions, for which the probability density function is known. For example, a user can compute the Fisher-Rao metric of the normal distributions with the syntax given below, which uses automatic differentiation behind the scenes.
\begin{minted}{python}
class MyInformationManifold(InformationManifoldMixin):
    def __init__(self):
        self.dim = 2
    def point_to_pdf(self, point):
        means = point[..., 0]
        stds = point[..., 1]
        def pdf(x):
            constant = (1. / gs.sqrt(2 * gs.pi * stds**2))
            return  constant * gs.exp(-((x - means) ** 2) / (2 * stds**2))
        return pdf
        
metric = FisherRaoMetric(
    information_manifold=MyInformationManifold(), support=(-10, 10))
\end{minted}
The user can then access the Fisher-Rao metric matrix $I(\theta)$ at $\theta = (1., 1.)$ with the code below.
\begin{minted}{python}
print(metric.metric_matrix(gs.array([1., 1.])))
>>> array([[1.00000000e+00, 1.11022302e-16],
       [1.11022302e-16, 2.00000000e+00]])
\end{minted}
We recognize here the metric matrix of the Fisher-Rao metric on the univariate normal distributions.
For convenience, the Fisher-Rao metrics for well-known parameter manifolds are already implemented in classes such as \codeobj{NormalMetric}, \codeobj{GammaMetric}, \codeobj{CategoricalMetric}, etc, as shown in Figure~\ref{fig:architecture}. These classes implement the closed-forms of the Fisher-Rao metric when these are known. The corresponding parameter manifolds in the classes \codeobj{NormalDistributions}, \codeobj{GammaDistributions}, \codeobj{CategoricalDistributions}, etc, are equipped with their Fisher-Rao metric, which is found as an attribute called \codeobj{metric}.

For example, the Fisher-Rao metric on the categorical distributions on a support of cardinal $5$ is found in the \codeobj{metric} attribute of the class of categorical distributions, i.e. \codeobj{metric = CategoricalDistributions(4).metric}. Its methods allow to compute exponential, logarithm maps and geodesics using \texttt{metric.exp()}, \texttt{metric.log()}, \texttt{metric.geodesic()}, together with the various notions of curvatures.

\section{Information manifolds implemented in Geomstats}\label{sec:catalogue}

This section details the tools of information geometry that we implement in each of the information manifold classes. As such, this section also provides a comprehensive review of the field of computational information geometry and its main applications. Each subsection further showcases code snippets using each information manifold to demonstrate the diversity of use cases of the proposed \codeobj{information\_manifold} module.

\subsection{One-dimensional parametric families}

\subsubsection{Main results}

\begin{figure}[h!]
    \centering
    \includegraphics[scale=.4]{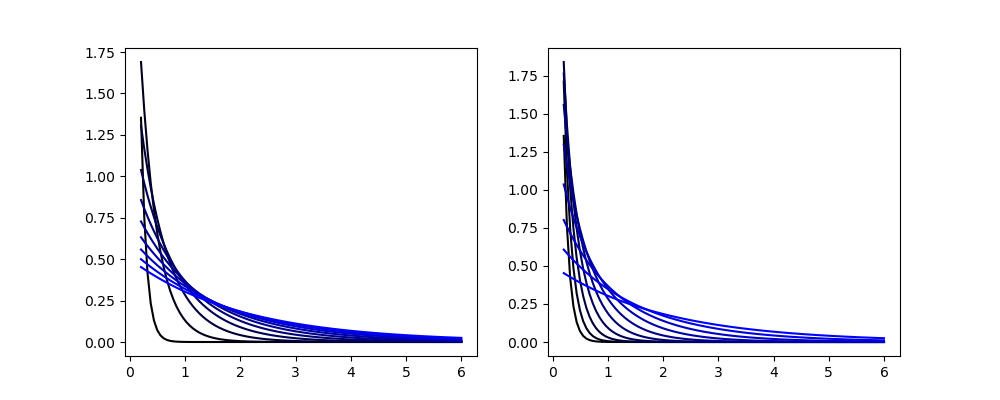}
    \caption{Comparison between affine (left) and geodesic (right) interpolations between pdfs of exponential distributions of parameter $\lambda_0=0.1$ (black) and $\lambda_1=2$ (blue).}
    \label{fig:1Dcomparison}
\end{figure}

The information geometry of one-dimensional information manifolds is simple: there is no curvature, the parameter manifold $\Theta$ is always diffeomorphic to $\R$, and there is only one path to go from one point to another in $\Theta$. However, the parametrization of this path can vary and leads to different interpolations between the probability distribution functions, as seen in Figure~\ref{fig:1Dcomparison}. 

The Fisher-Rao geodesic distances are given in closed forms for the Poisson, exponential, binomial (and Bernoulli) distributions in \cite{atkinson1981}. We compute it for geometric distributions too (see the appendix). Results are summarized in Table~\ref{tab:dim1} and implemented in the \codeobj{dist()} methods of the corresponding metric classes.

\begin{table}[h!]
\resizebox{\textwidth}{!}{%
\begin{tabular}{||c c c||} 
 \hline
 Distribution & P.d.f. (or P.m.f.) & Geodesic distance \\ [0.5ex] 
 \hline\hline
 Poisson (mean $\lambda$) & $\forall k \in \mathbb{N},\, P(k|\lambda) = \frac{\lambda^k}{k!} e^{-\lambda}, \, \lambda > 0$ & $d(\lambda_1, \lambda_2) = 2 |\sqrt{\lambda_1} - \sqrt{\lambda_2}|$ \\ [0.4ex]
 \hline
 Exponential (mean $\frac{1}{\lambda}$) & $\forall x \geq 0, f(x|\lambda) = \lambda e^{-\lambda x}, \, \lambda > 0$ & $d(\lambda_1, \lambda_2) = |\log \frac{\lambda_1}{\lambda_2}|$ \\ [0.4ex]
 \hline
 Binomial (known index $n$) & $\forall k \in \{0,...,n\},\, P(k|p) = \left(\begin{matrix} n \\ k \end{matrix} \right) p^k (1-p)^{n-k}, \, 0<p<1$ & $d(p_1, p_2) = 2 \sqrt{n} |\sin^{-1}(\sqrt{p_1}) - \sin^{-1}(\sqrt{p_2})|$ \\ [0.4ex]
 \hline
 Bernoulli ($1$-binomial) & $\forall k \in \{0,1\},\, P(k|p) =  p^k (1-p)^{1-k}, \, 0<p<1$ & $d(p_1, p_2) = 2 |\sin^{-1}(\sqrt{p_1}) - \sin^{-1}(\sqrt{p_2})|$ \\ [0.4ex]
 \hline
 Geometric & $\forall k \in \mathbb{N^*}, \, P(k|p) = (1-p)^{k-1}p, \, 0<p<1$ & $d(p_1, p_2) = 2|\tanh^{-1}(\sqrt{1-p_1}) - \tanh^{-1}(\sqrt{1-p_2}) |$ \\ \hline
 \end{tabular}}
 \smallskip
\caption{Fisher-Rao distance for one-dimensional parametric families of probability distributions implemented in the information geometry module. P.d.f. means probability density function and P.m.f. means probability mass function. These formulas are implemented in the \codeobj{dist()} methods in the metric Python classes of Figure~\ref{fig:architecture}.}
\label{tab:dim1}
\end{table}

\subsubsection{\codeobj{Geomstats} example}
The following code snippet shows how to compute the middle of the geodesic between points $p_1=.4$ and $p_2=.7$ on the one-dimensional $5$-binomial manifold.

\begin{minted}{python}
import geomstats.backend as gs
from geomstats.information_geometry.binomial import BinomialDistributions

manifold = BinomialDistributions(5)

point_a = .4
point_b = .7

times = gs.linspace(0, 1, 100)
geodesic = manifold.metric.geodesic(initial_point=point_a, end_point=point_b)(times)

middle = geodesic(.5)
print(middle)

>>>  0.5550055679356352

\end{minted}

The geodesic middle point of $p_1=.4$ and $p_2=.7$ on the $5$-binomial manifold is roughly $p=.555$, a little higher than the Euclidean middle point (=.55)!

\subsection{Multinomial and categorical distributions}


\subsubsection{Main results}

\textit{Multinomial distributions} model the results of an experiment with a finite number $k$ of outcomes, repeated $n$ times. When there is no repetition ($n=1$), it is called a \textit{categorical distribution}. Here the number of repetitions $n$ is always fixed. The parameter $\theta$ of the parameter manifold encodes the probabilities of the different outcomes. The parameter manifold $\Theta$ is therefore the interior of the $k-1$ dimensional simplex $\Theta =  \Delta_{k-1} = \{\theta \in \mathbb{R}^k: \forall i, \theta_i> 0, \theta_1+\hdots+\theta_k = 1 \}$.

\begin{definition}[Probability mass function of the multinomial distribution]
Given $k, n\in \mathbb{N^*}$ and  $\theta=(\theta_1, ..., \theta_k) \in \Delta_{k-1}$, the p.m.f. of the $n$-multinomial distribution of parameter $\theta$ is
$$p(x=(x_1,\hdots,x_k)|\theta) = \frac{n!}{x_1!\hdots x_k!} \theta_1^{x_1}\hdots\theta_k^{x_k},$$ 
where $x_i\in\{0,\hdots,n\}$ for all $i=1,\hdots,k$ and $x_1+\hdots+x_k=n$.
\end{definition}

The Fisher-Rao geometry on the parameter manifold $\Delta_{k-1}$ is well-known, see for example \cite{kass1989geometry}. We summarize the geometry with the following propositions.

\begin{proposition}[Fisher-Rao metric on the multinomial manifold]
The Fisher-Rao metric on the parameter manifold $\Theta = \Delta_{k-1}$ of $n$-multinomial distributions is given by
$$ds^2 = n\left(\frac{d\theta_1^2}{\theta_1} + ... + \frac{d\theta_k^2}{\theta_k} \right).$$
\end{proposition}



Thus, one can see that the Fisher-Rao metric on the parameter manifold $\Theta = \Delta_{k-1}$ of multinomial distributions can be obtained as the pullback of the Euclidean metric on the positive $(k-1)$-sphere of radius $2\sqrt{n}$, $S_{k-1}^+ = \{\theta \in \mathbb{R}^k: \forall i, \theta_i> 0, \sum_{i=1}^k \theta_i^2 = 2\sqrt{n} \}$ by the diffeomorphism
$$R : \theta \mapsto R(\theta) = (2\sqrt{n\theta_1}, ..., 2\sqrt{n\theta_k}).$$ 
Therefore the distance between two given parameters is the spherical distance of their images by transormation $R$, and the curvature of the parameter manifold is that of the $(k-1)$-sphere of radius $2\sqrt{n}$.
\begin{proposition}[Geodesic distance on the multinomial manifold]
The geodesic distance between two parameters $\theta^1, \theta^2 \in \Delta_{k-1}$ has the following analytic expression:
$$d(\theta^1, \theta^2) = 2\sqrt{n} \arccos{\left(\sum_{i=1}^k \sqrt{\theta_i^1 \theta_i^2} \right)}.$$
\end{proposition}

\begin{proposition}[Curvature of the multinomial manifold]
The Fisher-Rao manifold of multinomial distributions has constant sectional curvature $K=2\sqrt{n}$. 
\end{proposition}

\begin{figure}
    \centering
    \includegraphics[scale=.3]{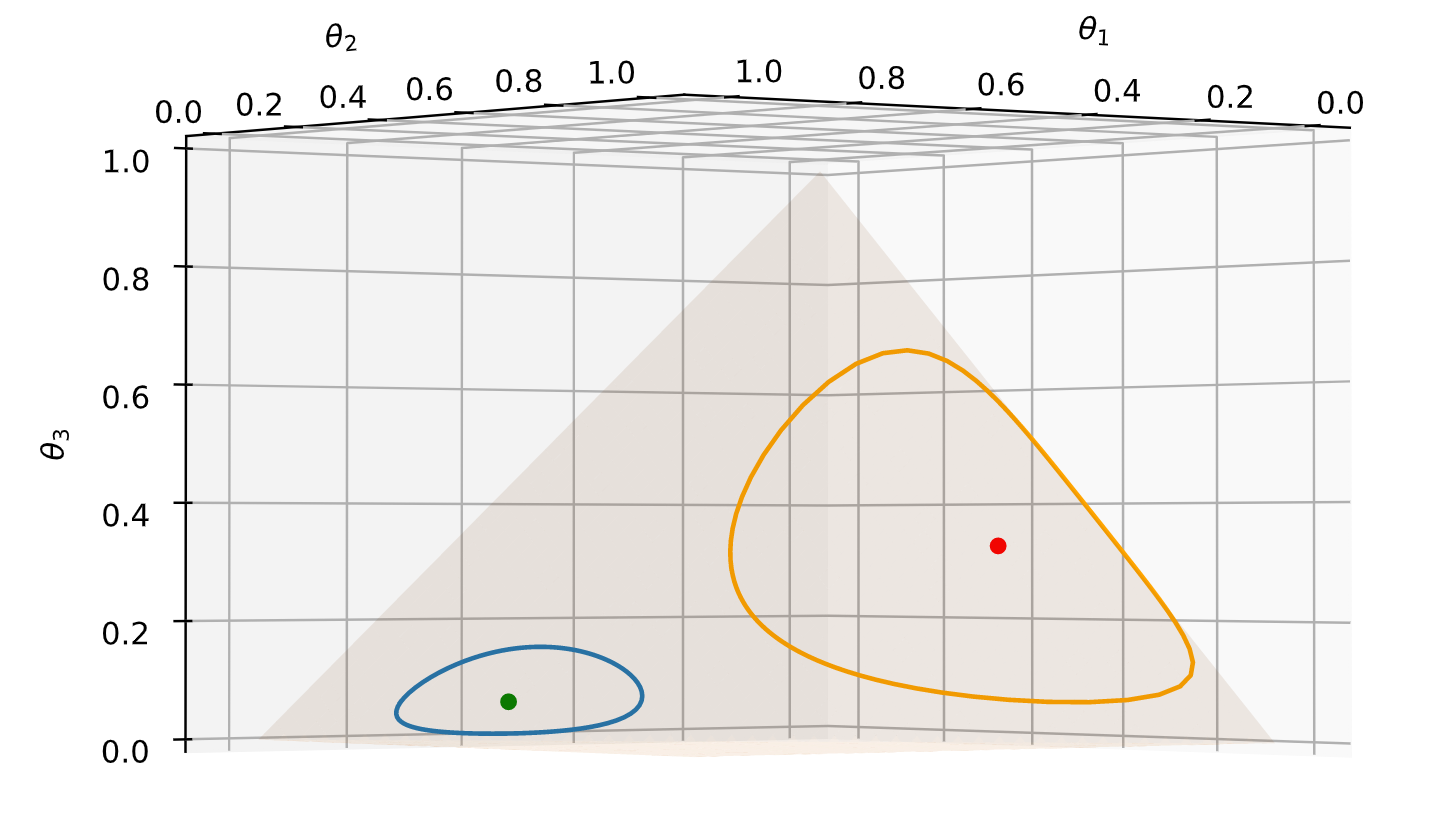}
    \captionsetup{singlelinecheck=off}
    \caption[]{Information geometry of the 3-Categorical manifold implemented in the Python class \codeobj{CategoricalDistributions}. The orange geodesic ball is of radius 0.7 and centered on the red point $(0.1, 0.58, 0.32)$, the blue geodesic ball is of radius 0.3 and centered on the green point $(0.74, 0.21, 0.05).$
    }
    \label{fig:categorical}
\end{figure}

We implement the p.m.f, Fisher-Rao metric, geodesic distance, and curvatures in the Python classes \codeobj{MultinomialDistributions} and \codeobj{MultinomialMetric} of the \codeobj{information\_geometry} module.

\subsubsection{Applications}

The Fisher-Rao geometry of multinomial distributions has been used in the literature, \textit{e.g.}, to formulate concepts in evolutionary game theory \cite{harper2009information} and to classify documents after term-frequency representation in the simplex \cite{lebanon2012learning}.

\subsubsection{\codeobj{Geomstats} example}

This example shows how we use the \codeobj{information\_geometry} module to compute on the $6$-categorical manifold, \textit{i.e.}, the $5$-dimensional manifold of categorical distributions with $k=6$ outcomes. The following code snippet computes the geodesic distances between a given point on the $6$-categorical manifold and the vertices of the simplex $\Delta_5$.

\begin{minted}{python}

import geomstats.backend as gs
from geomstats.information_geometry.categorical import CategoricalDistributions

manifold = CategoricalDistributions(dim=5)

point_a = gs.array([.1, .2, .1, .3, .15, .15])
point_b = gs.array([.25, .25, .1, .05, .05, .3])

vertices = list(gs.eye(6))

distances_a = [manifold.metric.dist(point_a, extremity) for vertex in vertices]
distances_b = [manifold.metric.dist(point_b, extremity) for vertex in vertices]

print(f"distances_a = {[float(str(distance)[:5]) for distance in distances_a]}")
print(f"distances_b = {[float(str(distance)[:5]) for distance in distances_b]}")

>>> distances_a = [2.498, 2.214, 2.498, 1.982, 2.346, 2.346]
>>> distances_b = [2.094, 2.094, 2.498, 2.69, 2.69, 1.982]

closest_a = vertices[gs.argmin(distances_a)]
closest_b = vertices[gs.argmin(distances_b)]

print(f"closest extremity to {point_a} is {closest_a}")
print(f"closest extremity to {point_b} is {closest_b}")

>>> closest extremity to [0.1  0.2  0.1  0.3  0.15 0.15] is [0. 0. 0. 1. 0. 0.]
>>> closest extremity to [0.25 0.25 0.1  0.05 0.05 0.3 ] is [0. 0. 0. 0. 0. 1.]

\end{minted}

This result confirms the intuition that the vertex of the simplex that is closest, in terms of the Fisher-Rao geodesic distance, to a given categorical distribution is the one corresponding to its mode. Indeed, noting $e_i = (\delta_{ij})_j$, $i=1,\hdots,6$ the extremities of the simplex, we see that for all $i\in\{1,\hdots, 6\}$ and $\theta \in \Delta_5$, $d(\theta, e_i) = \arccos({\sqrt{\theta_i}})$ is minimal when $i$ matches the mode of the distribution.







\subsection{Normal distributions}

Normal distributions are ubiquitous in probability theory and statistics, especially via the Central limit theorem. 
They are a very widely used modelling tool in practice, and provide one of the first non trivial Fisher-Rao geometries to be studied in the literature.

\subsubsection{Main results} Let us start by reviewing the univariate normal model.

\begin{definition}[Probability density function of the univariate normal distribution]
The p.d.f. of the normal distribution of mean $m\in\R$ and variance $\sigma^2\in\R_+^*$ is
$$f(x|\theta)=\frac{1}{\sqrt{2\pi\sigma^2}}\exp\left(-\frac{(x-m)^2}{2\sigma^2}\right).$$

\end{definition}

It is well known since the 1980s \cite{atkinson1981} that the corresponding Fisher-Rao metric with respect to $\theta=(m,\sigma)$ defines hyperbolic geometry on the parameter manifold $\Theta = \R\times\R_+^*$.

\begin{proposition}[Fisher-Rao metric for the univariate normal manifold]
The Fisher-Rao metric on the parameter manifold $\Theta = \mathbb{R} \times \mathbb{R_+^*}$ of normal distributions is
$$ds^2=\frac{dm^2+2d\sigma^2}{\sigma^2}.$$
\end{proposition}

Indeed, using the change of variables $m\mapsto m/\sqrt{2}$, we retrieve a multiple of the Poincaré metric $ds^2=2(dx^2+dy^2)/y^2$ on the upper half-plane $\{(x,y): \, x\in\R, y>0\}$, a model of two-dimensional hyperbolic geometry. Thus, closed-form expressions are known for the geodesics, which are either vertical segments or portions of half-circles orthogonal to the $m$-axis. The same is true for the distance. 

\begin{proposition}[Geodesic distance on the univariate normal manifold \cite{skovgaard1984}]
The geodesic distance between normal distributions of parameters $(m_1, \sigma_1)$ and $(m_2, \sigma_2)$ in $\mathbb{R} \times \mathbb{R_+^*}$ is given by
$$ d((m_1,\sigma_1),(m_2,\sigma_2))=\sqrt{2}\cosh^{-1}\left(
\frac{(m_1-m_2)^2/2 + (\sigma_1+\sigma_2)^2}{2\sigma_1\sigma_2}
\right).$$
\end{proposition}

The curvature is the same as that of the $2$-Poincaré metric, and rescaling the Poincaré metric by a factor $2$ implies dividing the sectional curvature by the same factor. The manifold of univariate normal distributions has therefore constant negative curvature, and since it is simply connected and geodesically complete we get the following result.
\begin{proposition}[Curvature of the univariate normal manifold \cite{skovgaard1984}.]
The Fisher-Rao manifold of normal distributions has constant sectional curvature $K=-1/2$. In particular, any two normal distributions can be linked by a unique geodesic, the length of which gives the Fisher-Rao distance. 
\end{proposition}

\begin{figure}[h!]
    \centering
    \centerline{
    \includegraphics[scale=0.5]{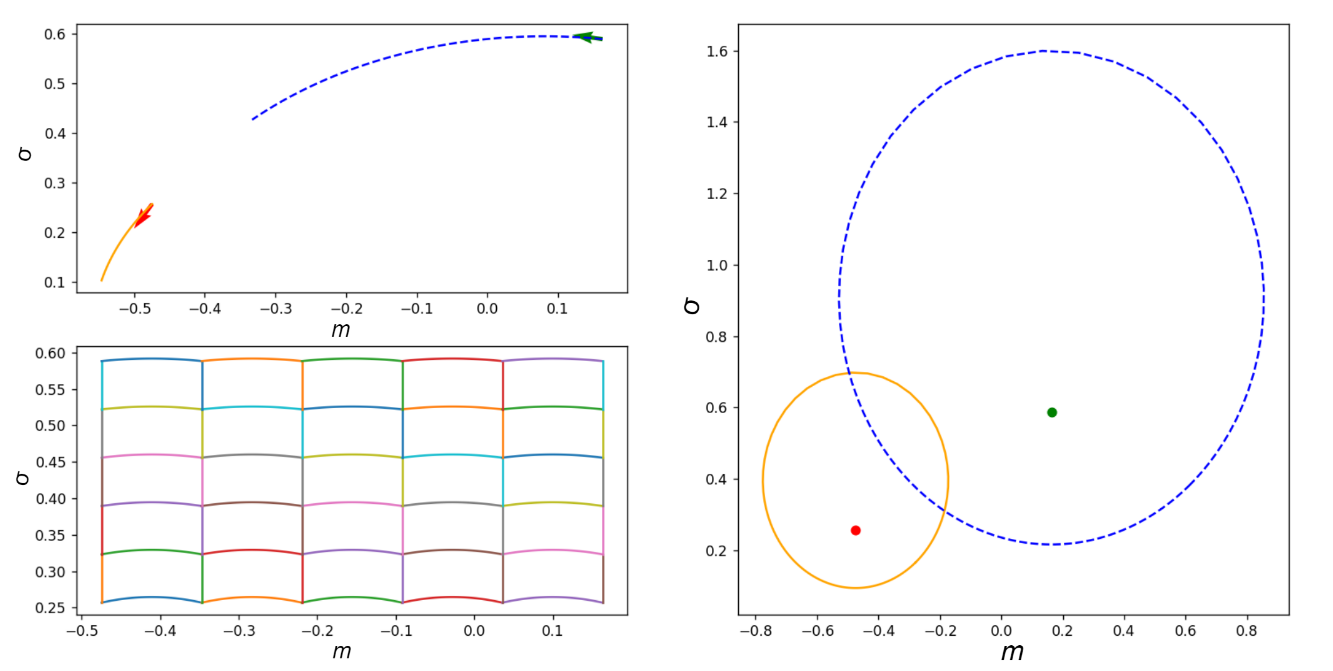}}
    \captionsetup{singlelinecheck=off}
    \caption[]{Information geometry of the manifold of normal distributions implemented in the Python class \codeobj{NormalDistributions}. Up-left: two geodesics of length 1 departing from two random points A and B;  Bottom-left: geodesic grid between A and B. Right: two geodesic spheres of radius 1 centered on A and B
    }
    \label{fig:univariate_normals}
\end{figure}

We implement the p.d.f, Fisher-Rao metric, geodesics, geodesic distance, and curvatures in the Python classes \codeobj{NormalDistributions} and \codeobj{NormalMetric} of the \codeobj{information\_geometry} module. Figure~\ref{fig:univariate_normals} shows 2 geodesics, 2 geodesic spheres, and 1 geodesic grid on the information manifold of univariate normal distributions.





We now turn to the multivariate case.
\begin{definition}[Probability density function of multivariate normal distributions]
In higher dimensions $p \geq 2$, the p.d.f. of the normal distribution of mean $m\in\R^p$ and covariance matrix $\Sigma \in S_p(\mathbb{R})^+$ is
$$f(x|\theta)=\frac{1}{\sqrt{(2\pi)p|\Sigma|}}\exp\left(-\frac{1}{2}(x-m)^\top\Sigma^{-1}(x-m)\right).$$
\end{definition}

The Fisher-Rao geometry of multivariate normal distributions was first studied in the early 1980's \cite{sato1979geometrical}, \cite{atkinson1981}\cite{skovgaard1984}. In general, no closed form expressions are known for the distance nor the geodesics associated to the Fisher information metric in the multivariate case. However, analytic expressions for these quantities are known for some particular submanifolds, and can be found e.g. in the review paper \cite{pinele2019fisher}.  The first of these particular cases corresponds to 
multivariate distributions with diagonal covariances.

\begin{proposition}[Multivariate normal distributions with diagonal covariance matrices \cite{skovgaard1984}]
The submanifold of Gaussian distributions with mean $m=(m_1,\hdots,m_p)$ and diagonal covariance matrix $\Sigma = diag(\sigma_1^2, ..., \sigma_p^2)$ can be identified with the product manifold $(\R\times\R_+^*)^p=\{(m_1,\sigma_1,\hdots, m_p,\sigma_p): \, m_i\in\R, \sigma_i>0\}$, on which the Fisher-Rao metric is the product metric
$$ds^2 = \sum_{i=1}^p \frac{dm_i^2 + 2d\sigma_i^2}{\sigma_i^2}.$$
The induced geodesic distance bertween distributions of means $m_j=(m_{ji})_{1\leq i\leq p}$ and covariance matrices $\Sigma_j=\mathrm{diag}(\sigma_{j1}^2,\hdots,\sigma_{jp}^2)$, $j=1, 2$, is given by
$$d_p((m_1,\Sigma_1),(m_2,\Sigma_2))=\sqrt{\sum_{i=1}^p d^2((m_{1i},\, \sigma_{1i}), (m_{2i},\sigma_{2i}))},$$
where $d$ is the geodesic distance on the space of univariate normal distributions.
\end{proposition}

The second particular case when the geometry is explicit corresponds to multivariate normal distributions with fixed mean. In this case, the parameter space is the space of symmetric positive definite matrices and the Fisher-Rao metric coincides with the affine-invariant metric \cite{pennec2006riemannian}. Note that even though the parameter with respect to which the Fisher information is computed differs between the different submanifolds of the multivariate normal distributions, this does not affect the distance, which is invariant with respect to diffeomorphic change of parametrization. 

\begin{proposition}[Multivariate normal distributions with fixed mean \cite{atkinson1981}]
Let $\mathbf{m} \in \mathbb{R}^p.$ The geodesic distance between Gaussian distributions with fixed mean $\mathbf{m}$ and covariance matrices $\Sigma_1$, $\Sigma_2$ is
$$d(\Sigma_1,\Sigma_2)=\sqrt{\frac{1}{2}\sum_{i=1}^p \log(\lambda_i)^2},$$
where the $\lambda_j$ are the eigenvalues of $\left(\Sigma_1\right)^{-1} \Sigma_2$.
\end{proposition}

The sectional curvature in the fixed mean case is negative, although non constant \cite{lenglet2006}. We implement the information geometry of the normal distributions reviewed here within the Python classes \codeobj{NormalDistributions} and \codeobj{NormalMetric} shown in Figure~\ref{fig:architecture}.

\subsubsection{Applications}

The Fisher-Rao geometry of normal distributions has proved very useful in the field of diffusion tensor imaging \cite{lenglet2006} and more generally in image analysis, \textit{e.g.}, for detection \cite{maybank2004}, mathematical morphology \cite{angulo2014morphological} and segmentation \cite{verdoolaege2011geodesics, strapasson2016clustering}. We refer the interested reader to the review paper \cite{pinele2019fisher} and the references therein.

\subsubsection{\codeobj{Geomstats} example}

This example shows how users can leverage the proposed \codeobj{information\_geometry} module to get intuition on the Fisher-Rao geometry of normal distributions. Specifically, we compute the geodesics and geodesic distance between two normal distributions with same variance and different means $m_1=1, m_2=4$, for two different values $\sigma^2=1, \sigma'^2=4$ of the common variance.

\begin{minted}{python}
import matplotlib.pyplot as plt
import geomstats.backend as gs
from geomstats.information_geometry.beta import BetaDistributions
from geomstats.information_geometry.normal import NormalDistributions

manifold = NormalDistributions()
point_a = gs.array([1., 1.])
point_b = gs.array([4., 1.])
point_c = gs.array([1., 2.])
point_d = gs.array([4., 2.])

print(manifold.metric.dist(point_a, point_b))
print(manifold.metric.dist(point_c, point_d))

>>> 2.38952643457422
>>> 1.3862943611198915

times = gs.linspace(0, 1, 100)
geod_ab = manifold.metric.geodesic(initial_point=point_a, end_point=point_b)(times)
geod_cd = manifold.metric.geodesic(initial_point=point_c, end_point=point_d)(times)

max_variance_ab = geodesic_ab[gs.argmax(geod_ab[:, 1])]
max_variance_cd = geodesic_cd[gs.argmax(geod_cd[:, 1])]

plt.plot(*gs.transpose(geod_ab))
plt.scatter(*point_a, color='g')
plt.scatter(*point_b, color='g')
plt.scatter(*max_variance_ab, color='r')
plt.plot(*gs.transpose(geod_cd))
plt.scatter(*point_c, color='g')
plt.scatter(*point_d, color='g')
plt.scatter(*max_variance_cd, color='r')
plt.ylim([0., 3.])
plt.show()

\end{minted}

\begin{figure}
    \includegraphics[width=0.5\linewidth]{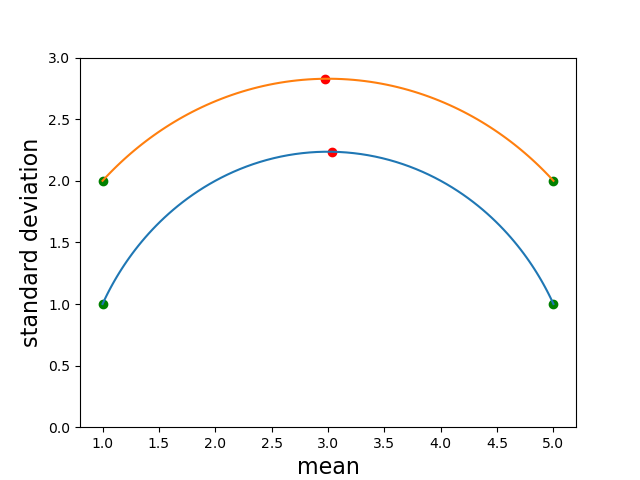}
    \caption{Geodesics in the manifold of normal distributions. When the variance of the normal distributions at the extremities (green points) increases, the geodesic becomes shorter. Variance increases along the geodesic and reaches a maximum in the middle (red points).}
    \label{fig:normal_geod}
\end{figure}

The two geodesics generated by this code snippet yield the two curves in Figure~\ref{fig:normal_geod}. We see that the higher the variance, the smaller the distance. As pointed out in \cite{costa2015fisher}, this result reflects the fact that the p.d.f.s overlap more when the variance increases. On each geodesic, we observe that the point of maximum variance corresponds to the geodesic' middle point.

%
%
%
%
%
%

\subsection{Gamma distributions}

Gamma distributions form a $2$-parameter family of distributions defined on the positive half-line, and are used to model the time between independent events that occur at a constant average rate. They have been widely used to model right-skewed data, such as cancer rates \cite{shinmoto2015diffusion}, insurance claims \cite{semenikhine2018multiplicative}, and rainfall \cite{husak2007use}. 


\subsubsection{Main results}

Standard Gamma distributions take support over $\mathbb{R_+^*}$ and consist of one of the prime examples of information geometry, namely for for the variety of parametrizations they have been endowed with \cite{lauritzen1987statistical}, \cite{burbea2002some}, \cite{arwini2008}.

\begin{definition}[Probability density function for Gamma distributions in natural coordinates]
In natural coordinates, given $(\nu, \kappa) \in \left(\mathbb{R_+^*}\right)^2$, the p.d.f. of the two-parameter Gamma distribution of rate $\nu$ and shape $\kappa$ is:
$$\forall x > 0,\, f(x | \nu, \kappa) = \frac{\nu^{\kappa}}{\Gamma(\kappa)} x^{\kappa-1} e^{-\nu x} \text{, where $\Gamma$ is the Gamma function}.$$
\end{definition}

\begin{proposition}[Fisher-Rao metric for the Gamma manifold in natural coordinates \cite{arwini2008}]
The Fisher-Rao metric on the Gamma manifold $\Theta = \left(\mathbb{R_+^*}\right)^2$ is $$ds^2 = \frac{\kappa}{\nu^2} d\nu^2 - 2\frac{d\nu d\kappa}{\nu} + \psi'(\kappa) d\kappa^2 ,$$ where $\psi$ is the digamma function, i.e. $\psi = \frac{\Gamma'}{\Gamma}$.
\end{proposition}
However, the fact that this metric is not diagonal for the natural parametrization encourages one to consider the manifold under a different set of coordinates. Getting rid of the middle term in $d\nu d\kappa$ highly simplifies the geometry.

\begin{proposition}[Fisher-Rao metric for the Gamma manifold in $(\gamma, \kappa)$ coordinates \cite{arwini2008}]
The change of variable $(\gamma, \kappa) = (\frac{\kappa}{\nu}, \kappa)$ gives the following expression of the Fisher-Rao metric: 
$$ds^2 = \frac{\kappa}{\gamma^2} d\gamma^2 + \left(\psi'(\kappa) - \frac{1}{\kappa}\right) d\kappa^2.$$
\end{proposition}

Both parametrizations $(\gamma, \kappa)$ and $(\kappa, \gamma)$ can be found in the literature. The use of  of $(\kappa, \gamma)$ is standard in information geometry and it is the one we use to implement the Gamma manifold. This yields the following expression of the p.d.f.

\begin{definition}[Probability density function for Gamma distributions in $(\kappa, \gamma)$  coordinates]
The p.d.f. of the two-parameter Gamma distribution of parameters $\gamma$, $\kappa$ is:
$$\forall x > 0,\, f(x | \gamma, \kappa) = \frac{\kappa^{\kappa}}{\gamma^\kappa\Gamma(\kappa)} x^{\kappa-1} e^{-\frac{\kappa x}{\gamma}}.$$
\end{definition}


\begin{proposition}[Geodesic equations on the Gamma manifold \cite{arwini2008}]
The associated geodesic equations are:
$$
\begin{cases}
    \ddot{\gamma} = \frac{\dot{\gamma}^2}{\gamma} - \frac{\dot{\gamma} \dot{\kappa}}{\kappa} \\
    \ddot{\kappa} = \frac{\kappa \dot{\gamma}^2}{2 \gamma^2 (\kappa \psi'(\kappa)-1)} - \frac{(\psi"(\kappa)\kappa^2 + 1) \dot{\kappa}^2}{2 \kappa (\kappa \psi'(\kappa)-1)}.
\end{cases}
$$
\end{proposition}

\begin{figure}[h!]
    \centering
    \centerline{
    \includegraphics[scale=.4]{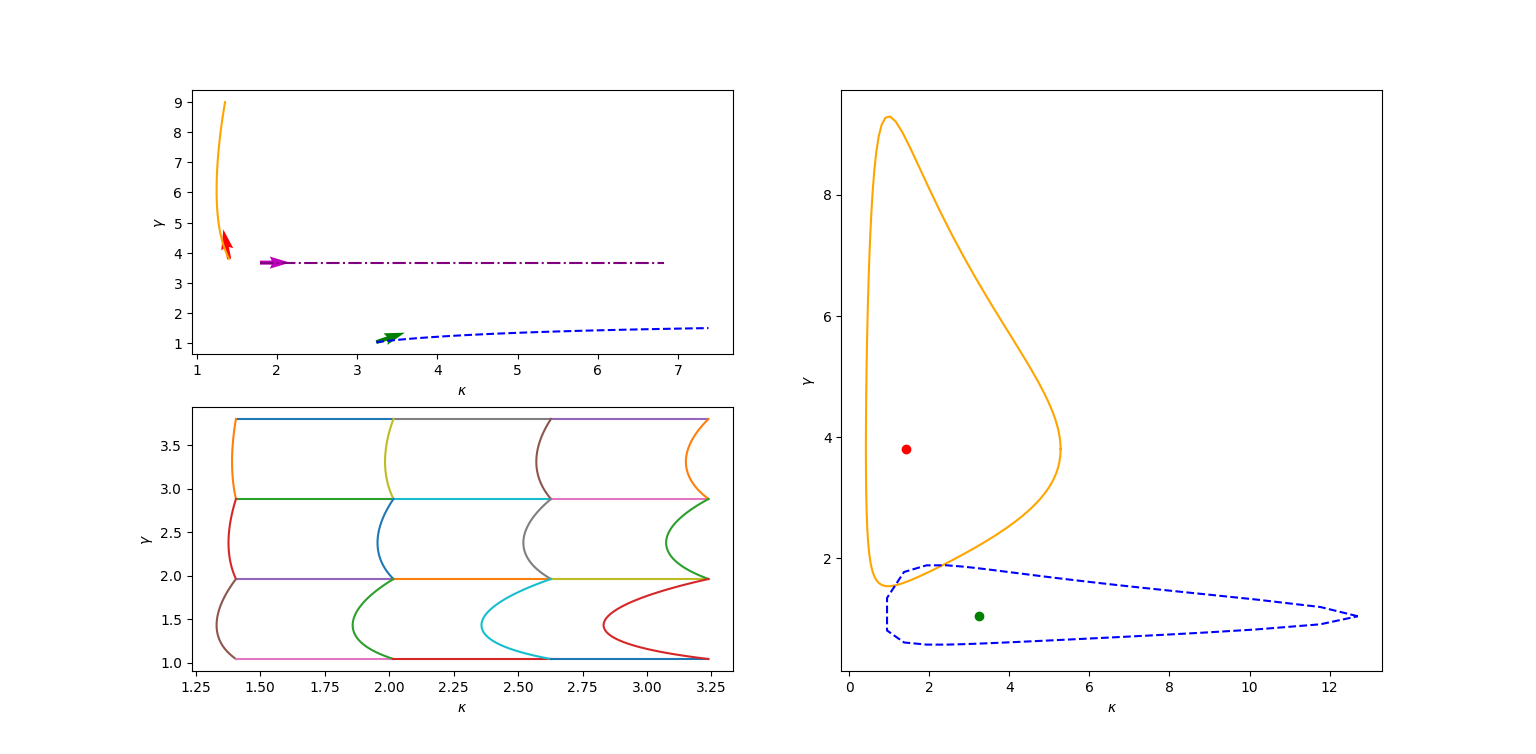}}
    \captionsetup{singlelinecheck=off}
    \caption[]{Information geometry of the manifold of Gamma distributions implemented in the Python class \codeobj{GammaDistributions}. Up-left: three geodesics of length 1 departing from two random points A (red) and B (green) and C (magenta, with $\gamma$ constant). Bottom-left: geodesic grid between A and B. Right: two geodesic spheres of radius 1 centered on A and B; 
    }
    \label{fig:gamma_geodesics}
\end{figure}

No closed form expressions are known for the distance nor the geodesics associated to the Fisher information geometry with respect to  $(\gamma, \kappa)$. Yet, our information module is able to compute both numerically by leveraging the automatic differentiation computations available in the parent Python class of the \codeobj{FisherRaoMetric}. Figure~\ref{fig:gamma_geodesics} shows 3 geodesics, 2 geodesic spheres, and a geodesic grid for the Gamma manifold. Running code from the information geometry module shows that some geodesics are horizontal (with $\gamma$ constant), which is notable. This can also be directly seen from the geodesic equation $\ddot{\gamma} =  \dot{\gamma} \left(\frac{\dot{\gamma}}{\gamma} - \frac{ \dot{\kappa}}{\kappa} \right)$: a geodesic with a horizontal initial direction ($\dot\gamma = 0$) will stay horizontal.


There is a closed-form expression of the geodesic distance in the manifold of Gamma distributions with fixed $\kappa$, which is therefore a one-dimensional manifold.

\begin{proposition}[Geodesic distance on the Gamma manifold with fixed $\kappa$.] The geodesic distance $d$ on the Gamma manifold, for a fixed $\kappa$ is given in $(\kappa, \gamma)$ parameterization by:
 $$\forall \gamma_1, \gamma_2 >0,\, d(\gamma_1, \gamma_2) = \sqrt{\kappa}  \left|\log \frac{\gamma_1}{\gamma_2}\right|,$$
 or, in $(\kappa, \nu)$ parameterization by:
 $$\forall \gamma_1, \gamma_2 >0,\, d(\nu_1, \nu_2) = \sqrt{\kappa} \left|\log \frac{\nu_1}{\nu_2} \right|.$$
\end{proposition}

This result, proved in the appendix, was expected, at least for integer values of $\kappa$. Consider one Gamma process as the sum of $\kappa$ i.i.d exponential processes. Because the processes are independent, the Fisher information for the Gamma distribution is $\kappa$ times as big as that of the exponential distribution. Consequently, the length of a geodesic on the Gamma manifold observes a $\sqrt{\kappa}$ coefficient.

\begin{figure}[h!]
    \centering
    \includegraphics[scale=.5]{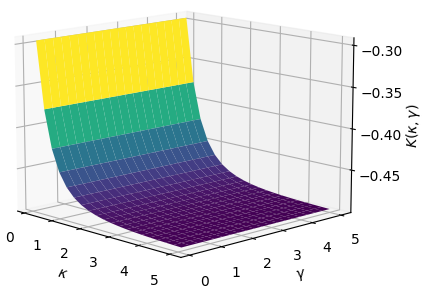}
    \includegraphics[scale=.5]{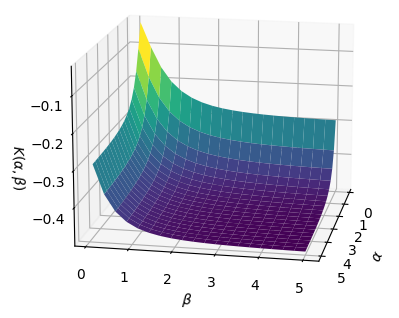}
    \caption{Sectional curvature of the Fisher-Rao manifolds of gamma (left) and beta (right) distributions.}
    \label{fig:gamma_curvature}
\end{figure}

The sectional curvature of the Gamma manifold, which is plotted in Figure~\ref{fig:gamma_curvature}, is everywhere negative, bounded and depends only on the $\kappa$ parameter. Since it is also simply connected and geodesically complete, the following result holds.

\begin{proposition}[Curvature of the Gamma manifold \cite{burbea2002some}]
The sectional curvature of the Gamma manifold at each point $(\gamma,\, \kappa) \in \left(\mathbb{R_+^*}\right)^2$ verifies
$$-\frac{1}{2} < K(\gamma,\kappa)=K(\kappa)=\frac{\psi'(\kappa) + \kappa \psi''(\kappa)}{4(-1 + \kappa \psi'(\kappa))^2} < -\frac{1}{4}.$$ 
In particular, any two gamma distributions can be linked by a unique geodesic in the parameter space, the length of which gives the Fisher-Rao distance.
\end{proposition}

We implement the information geometry of the Gamma distributions reviewed here within the Python classes \codeobj{GammaDistributions} and \codeobj{GammaMetric} shown in Figure~\ref{fig:architecture}. Let us mention that the Fisher-Rao geometry of generalized Gamma distributions have also been studied in the literature \cite{chen2013riemannian, abbad2017rao, rebbah2019}, and will be the object of future implementation in the proposed information geometry module.

\subsubsection{Applications}

Information geometry of both the standard Gamma and the generalized Gamma manifolds have been used in the literature. Most often, the goal is to implement a ``natural'' (geodesic) distance between distributions. In that aspect, a geometric reasoning of Gamma distributions finds purposes in many fields, ranging from performance improvement in classification methods in medical imaging \cite{rebbah2019} to texture retrieval \cite{abbad2017rao}.

\subsubsection{\codeobj{Geomstats} example} 
In the following example, we compute the sectional curvature of the Gamma manifold at a given point. The sectional curvature is computed for the subspace spanned by two tangent vectors, but since the gamma manifold is two dimensional, the result does not depend on the chosen vectors.

\begin{minted}{python}
import geomstats.backend as gs
from geomstats.information_geometry import GammaDistributions

dim = 2
manifold = GammaDistributions()
point = gs.array([1., 2.])

vec_a = manifold.to_tangent(gs.random.rand(dim)))
vec_b = manifold.to_tangent(gs.random.rand(dim)))
vec_c = manifold.to_tangent(gs.random.rand(dim)))

print(manifold.metric.curvature(vec_a, vec_b, point))
print(manifold.metric.curvature(vec_a, vec_c, point))

>>> -0.45630369144018423
>>> -0.4563036914401915
\end{minted}

A comprehensive example using information geometry of the Gamma manifold in the context of traffic optimization in São Paulo can be found in \href{https://notebooks.gesis.org/binder/jupyter/user/geomstats-geomstats-yo5d6iiw/notebooks/notebooks/18_real_world_applications__sao_paulo_traffic_optimization.ipynb}{this notebook}.

\subsection{Beta and Dirichlet distributions}

\subsubsection{Main results}

Beta distributions form a 2-parameter family of probability measures defined on the unit interval and often used to define a probability distribution on probabilities. In Bayesian statistics, it is the conjugate prior to the binomial distribution, meaning that if the prior on the probability of success in a binomial experiment belongs to the family of beta distributions, then so does the posterior distribution. This allows users to estimate the distribution of the probability of success by iteratively updating the parameters of the beta prior. Beta and Dirichlet distributions are defined as follows:

\begin{definition}[Probability density function of Beta distributions]
The p.d.f. of beta distributions is parameterized by two shape parameters $\alpha,\beta>0$ and given by:
$$f(x|\alpha, \beta)=\frac{\Gamma(\alpha+\beta)}{\Gamma(\alpha)\Gamma(\beta)}x^{\alpha-1}(1-x)^{\beta-1}, \quad \forall x\in[0, 1].$$
\end{definition}

\begin{figure}
    \centering
    \includegraphics[scale=.4]{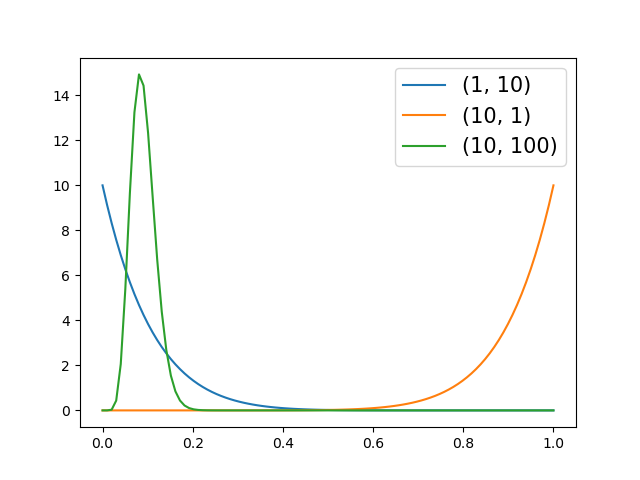}
    \caption{P.d.f.s of the beta distributions plotted in Geomstats example \ref{ex:beta}. The Fisher-Rao geodesic distance between the parameters of the blue and green distributions is larger than the one between the blue and orange, while the converse is true for the Euclidean distance.}
    \label{fig:beta_manifold}
\end{figure}

Figure~\ref{fig:beta_manifold} shows examples of p.d.f. of beta distributions, which can take a wide variety of shapes. The distribution has a unique mode in $]0,1[$ when $\alpha,\beta>1$, and a mode in $0$ or $1$ otherwise.

Beta distributions can be seen as a sub-family of the Dirichlet distributions, defined on the $(n-1)$-dimensional probability simplex 
$\Delta_{n-1}$ of $n$-tuples composed of non-negative components that sum up to one. Similarly to the beta distribution, the Dirichlet distribution is used in Bayesian statistics as the conjugate prior to themultinomial distribution. It is a multivariate generalization of the beta distribution in the sense that if $X$ is a random variable following a beta distribution of parameters $\alpha_1,\alpha_2$, then $(X,1-X)$ follows a Dirichlet distribution of same parameters on $\Delta_1$. 

\begin{definition}[Probability density function of Dirichlet distributions]
The p.d.f. of Dirichlet distributions is parametrized by $n$ positive reals $\alpha_1,\hdots,\alpha_n>0$ and given by:
$$f(x|\alpha_1,\hdots,\alpha_n)=\frac{\Gamma(\sum_{i=1}^n\alpha_i)}{\prod_{i=1}^n\Gamma(\alpha_i)}\prod_{i=1}^n{x_i}^{\alpha_i-1}, \quad \forall (x_1,\hdots,x_n)\in\Delta_{n-1}.$$
\end{definition}

\begin{proposition}[The Fisher-Rao metric on the Dirichlet manifold \cite{lebrigant2021fisher}]
The Fisher-Rao metric on the parameter manifold $\Theta=(\mathbb R_+^*)^n$ of Dirichlet distributions is
$$ds^2=\sum_{i=1}^n\psi'(\alpha_i)d\alpha_i^2-\psi'(\bar\alpha)d\bar\alpha^2,$$
where $\bar\alpha=\sum_{i=1}^n\alpha_i$.
\end{proposition}

\begin{figure}[h!]
    \centering
    \centerline{
    \includegraphics[scale=.4]{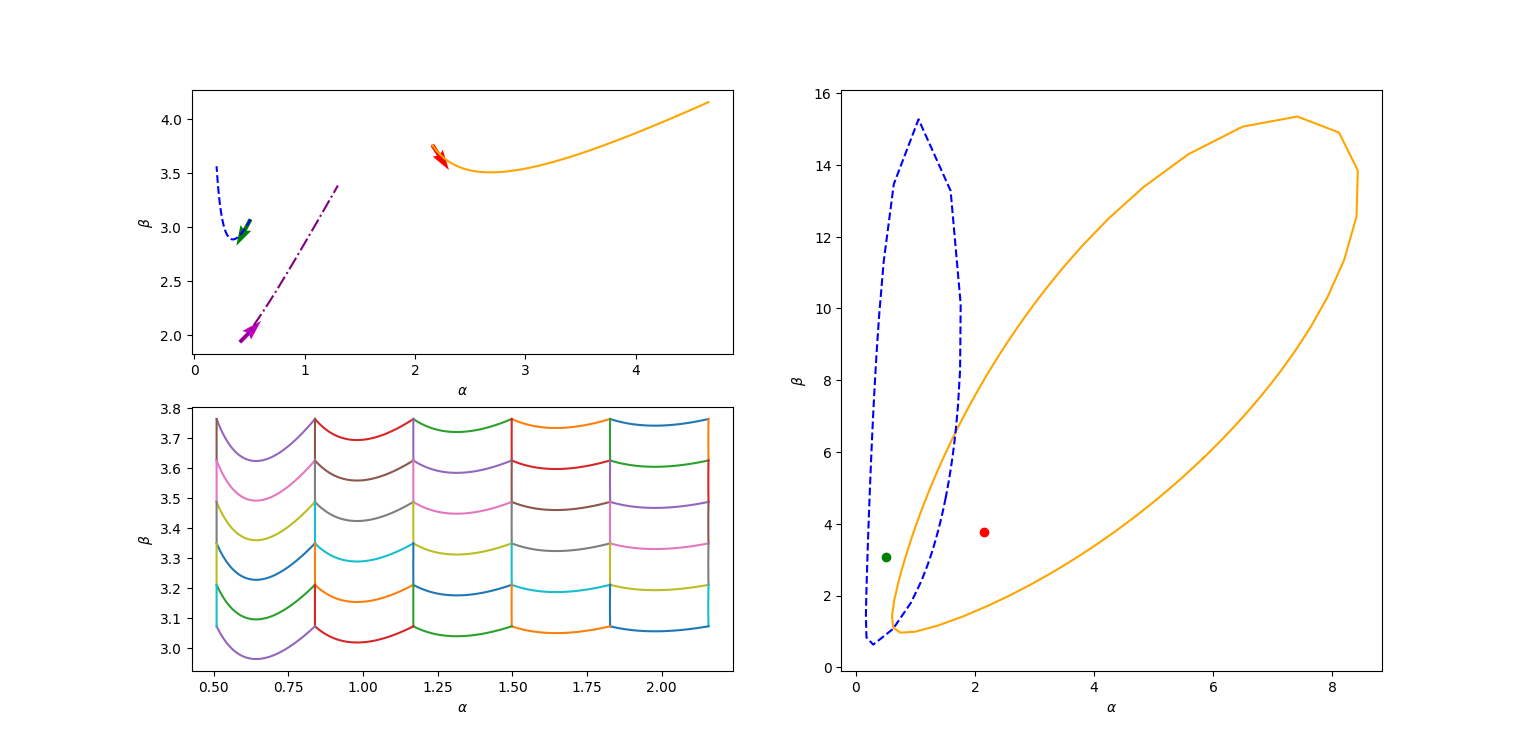}}
    \captionsetup{singlelinecheck=off}
    \caption[]{Information geometry of the Beta manifold implemented in \codeobj{BetaDistributions}. Up-left: three geodesics of length 1 departing from three random points A (red) and B (green) and C (magenta, with $\frac{\alpha}{\beta}$ constant). Bottom-left: geodesic grid between A and B. Right: two geodesic spheres of unit radius centered on A and B; 
    }
    \label{fig:beta_geodesic}
\end{figure}

\begin{figure}[h!]
    \centering
    \raisebox{.4cm}{
    \includegraphics[scale=.38]{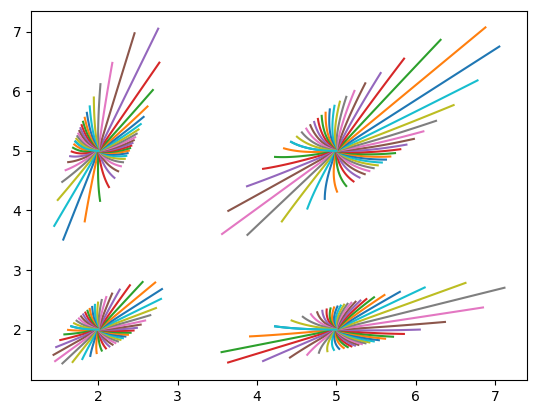}}
    \includegraphics[scale=.4]{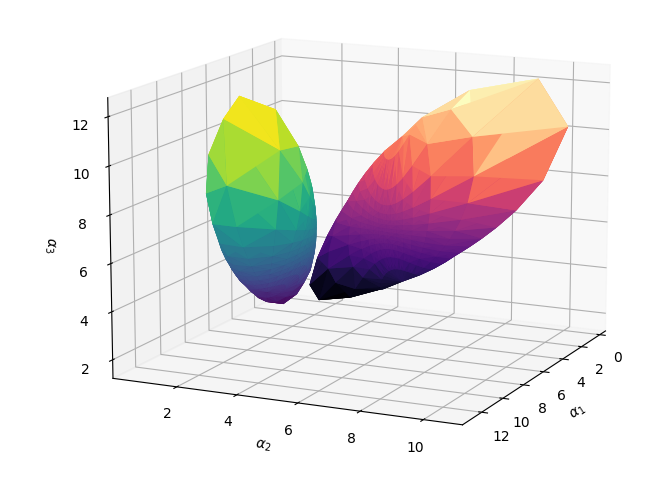}
    \caption{Left: rays of four geodesic spheres in the beta manifold, the oval shape of which suggest that the cost to go from one beta distribution to another is less important along the lines of equation $\alpha_2/\alpha_1=\text{cst}$. This seems natural since these are the lines of constant distribution mean. Right: geodesic spheres of unit radius in the 3-Dirichlet manifold. 
    }
    \label{fig:dirichlet_geodesic}
\end{figure}

No closed form are known for the geodesics of the beta and Dirichlet manifold. Therefore, our \codeobj{information\_geometry} module solves the geodesic equations numerically. Figure~\ref{fig:beta_geodesic} shows 3 geodesics, 2 geodesic sphere and 1 geodesic grid for the beta manifold, and Figure-\ref{fig:dirichlet_geodesic} shows geodesic spheres in the $3$-Dirichlet manifold. In the beta manifold, the oval shape of the geodesic spheres  suggest that the cost to go from one point to another is less important along the lines of equation $\alpha_2/\alpha_1=\text{cst}$. This seems natural since these are the lines of constant distribution mean.

The Dirichlet manifold is isometric to a hypersurface in flat $(n+1)$-dimensional Minkowski space through the transformation $$(x_1,\hdots,x_n)\mapsto (\eta(x_1),\hdots,\eta(x_n),\eta(x_1+\hdots+x_n)),$$
where $\eta'(x)=\sqrt{\psi'(x)}$. This allows to show the following result on the curvature, which is plotted in Figure~\ref{fig:gamma_curvature} for dimension $2$.

\begin{proposition}[\cite{lebrigant2021fisher}]
The parameter manifold of Dirichlet distributions endowed with the Fisher-Rao metric is simply connected, geodesically complete and has everywhere negative sectional curvature. In particular, any two Dirichlet distributions can be linked by a unique geodesic, the length of which gives the Fisher-Rao distance.
\end{proposition}


\begin{figure}
    \centering
    \includegraphics[trim=100 0 100 0, scale=.4]{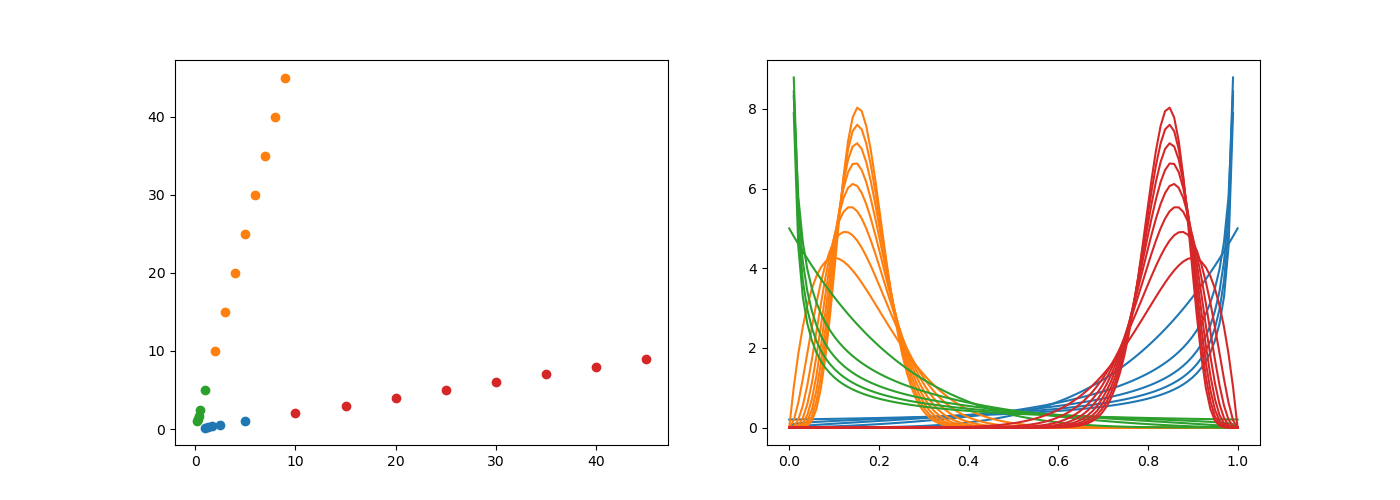}
    \includegraphics[trim=100 0 100 20, scale=.4]{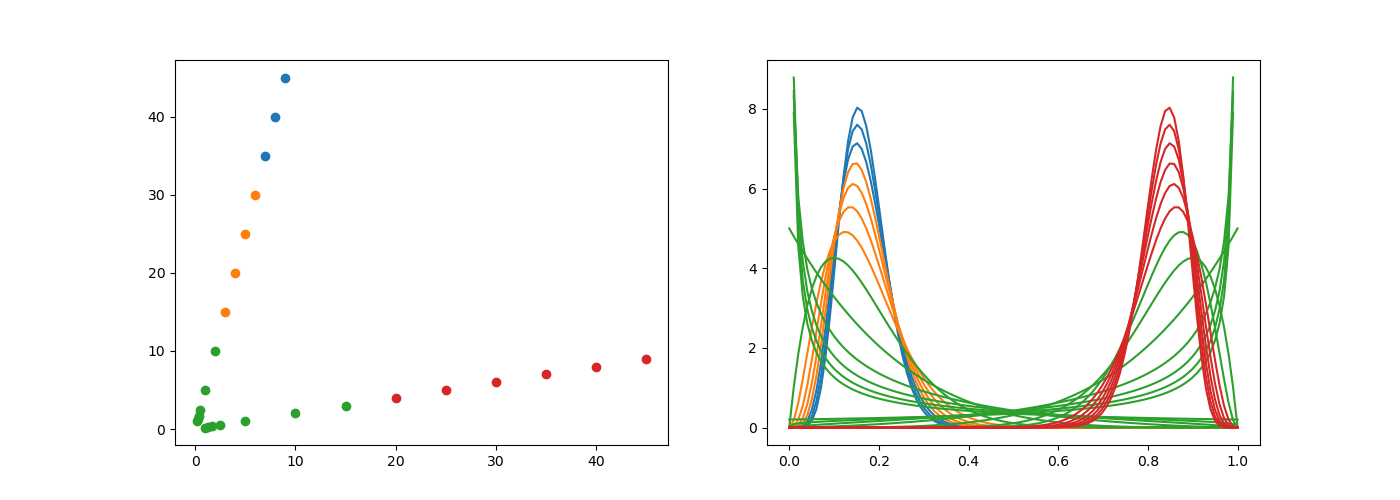}
    \caption{Results of K-means clustering of the beta distributions of Geomstats example \ref{ex:beta} using the Fisher-Rao metric (upper row) and the Euclidean distance (lower row), shown in terms of parameters (left column) and p.d.f.s (right column). Contrary to the Euclidean distance, the Fisher-Rao metric regroups the distributions with the same mean, i.e. with parameters aligned on a straight line through the origin, and inside a group of same mean, it regroups the p.d.f.s with similar shape.} 
    \label{fig:beta_clustering}
\end{figure}

The classes \codeobj{BetaDistributions}, \codeobj{DirichletDistributions}, and \codeobj{DirichletMetric} implement the geometries described here. We note that \codeobj{BetaDistributions} inherits from \codeobj{DirichletDistributions} and thus inherits the computations coming from its Fisher-Rao metrics as shown in Figure~\ref{fig:architecture}.

\subsubsection{Applications} The Fisher-Rao geometry of beta distributions has received less attention in the literature than the previously described families, although it has been used in \cite{lebrigant2021classifying} to classify histograms of medical data.

\subsubsection{\codeobj{Geomstats} example}\label{ex:beta} 





The following example compares the Fisher-Rao distance with the Euclidean distance between the beta distributions shown in Figure~\ref{fig:beta_manifold}. The Euclidean distance between the beta distributions with p.d.f.s shown in blue and green is much larger than the one between the blue and orange. This does not seem satisfactory when considering the differences in mean and mass overlap. By contrast, the blue distribution is closer to the green than to the orange distribution according to the Fisher-Rao metric.

\begin{minted}{python}
import matplotlib.pyplot as plt

import geomstats.backend as gs
from geomstats.information_geometry.beta import BetaDistributions

point_a = gs.array([1., 10.])
point_b = gs.array([10., 1.])
point_c = gs.array([10., 100.])

# Plot pdfs
samples = gs.linspace(0., 1., 100)
points = gs.stack([point_a, point_b, point_c])
pdfs = manifold.point_to_pdf(points)(samples)
plt.plot(samples, pdfs)
plt.show()

# Euclidean distances
print(gs.linalg.norm(point_a - point_b))
print(gs.linalg.norm(point_a - point_c))

>>> 12.73
>>> 90.45

# Fisher-Rao distances
print(manifold.metric.dist(point_a, point_b))
print(manifold.metric.dist(point_a, point_c))

>>> 4.16
>>> 1.76
\end{minted}

More generally, beta distributions with the same mean are close for the Fisher-Rao metric. Indeed, the oval shape of the geodesic balls shown in Figure~\ref{fig:beta_manifold} suggests that the cost to go from one point to another is less important along the lines of equation $\alpha_2/\alpha_1=\text{cst}$, which are the lines of constant distribution mean. 

The next example performs K-means clustering , using either the Euclidean distance or the Fisher-Rao distance. We consider a set of beta distributions whose means take only two distinct values, which translates into the alignment of the parameters on two straight lines going through the origin, see Figure~\ref{fig:beta_clustering}. The clustering based on the Fisher-Rao metric (top row of the figure) distinguishes these two classes, and can further separate the distributions according to the shape of their p.d.f. The Euclidean distance on the other hand (bottom row of the figure) does not distinguish between the two different means.

\begin{minted}{python}
import geomstats.backend as gs
from geomstats.geometry.euclidean import Euclidean
from geomstats.information_geometry.beta import BetaDistributions
from geomstats.learning.kmeans import RiemannianKMeans

# Data
values = gs.array([1/i for i in range(1, 6)] + [i for i in range(2, 10)])

factor = 5
cluster_1 = gs.stack((values, factor * values)).T
cluster_2 = gs.stack((factor * values, values)).T

points = gs.vstack((cluster_1, cluster_2))

n_points = points.shape[0]
n_clusters = 4

# KMeans with the Euclidean distance
r2 = Euclidean(dim=2)

kmeans = RiemannianKMeans(metric=r2.metric, n_clusters=n_clusters, verbose=1)
centroids_eucl = kmeans.fit(points)
labels_eucl = kmeans.predict(points)

# KMeans with the Fisher Rao distance
beta = BetaDistributions()

kmeans = RiemannianKMeans(metric=beta.metric, n_clusters=n_clusters, verbose=1)
centroids_riem = kmeans.fit(points)
labels_riem = kmeans.predict(points)
\end{minted}


\section{Application to text classification}\label{sec:application}

This section presents a comprehensive usecase of the proposed Geomstats module \codeobj{information\_geometry} for text classification using the information manifold of Dirichlet distributions.

We use the Latent Dirichlet Allocation (LDA) model to represent documents in the parameter manifold of Dirichlet distributions. LDA is a generative model for text, where each document is seen as a random mixture of topics, and each topic as a categorical distribution over words \cite{blei2003latent}. Specifically, consider a corpus with several documents composed of words from a dictionary of size $V$, and $K$ topics represented by a $K\times V$ matrix $\beta$ where the $i$-th line $\beta_{i\bullet}$ gives the discrete probability distribution of the $i$-th topic over the vocabulary. Given a Dirichlet parameter $\alpha$ in $\Delta_{K-1}$ the $(K-1)$-dimensional simplex, each document of $N$ words is generated as follows. First, we sample mixing coefficients $\theta=(\theta_1,\hdots,\theta_K) \sim \mathrm{Dirichlet}(\alpha)$. Next, in order to generate each word, we sample the $i$-th topic from $\mathrm{Categorical}(\theta)$. Finally, we sample a word from $\mathrm{Categorical}(\beta_{i\bullet})$. In other words, for each document the following two steps are iterated for $n=1,\hdots,N$:
\begin{enumerate}
\item select a topic $z_n$ according to $\mathbb P(z_n=i | \theta)=\theta_i$, $1\leq i\leq K$
\item select a word $w_n$ according to $\mathbb P(w_n=j|z_n,\beta)=\beta_{ij}$, $1\leq j\leq V$.
\end{enumerate}
Here ``$z_n=i$'' means that the $i^{th}$ topic is selected among the $K$ possible topics, and this is encoded as a vector of size $K$ full of zeros except for a $1$ in $i^{th}$ position. Similarly, ``$w_n=j$'' means that the $j^{th}$ word of the dictionary is selected and is encoded by a vector of size $V$ full of zeros except for a $1$ in $j^{th}$ position. 

The Dirichlet parameter $\alpha\in\Delta_{K-1}$ and the word-topic distributions $\beta \in \mathbb R^{k\times V}$ are the parameters of the model, which need to be estimated from data. Unfortunately, the likelihood of the LDA model cannot be computed and therefore cannot be maximized directly to estimate these parameters. In the seminal paper \cite{blei2003latent}, the authors introduce variational parameters that are document-specific, as well as a lower bound of the likelihood that involves these parameters. This bound can serve as a substitute for the true likelihood when estimating the parameters $\alpha$ and $\beta$. In binary classification experiments, the authors use the variational Dirichlet parameters to represent documents of the Reuters-21578 dataset and perform Euclidean support vector machine (SVM) in this low-dimensional representation space. 

\begin{figure}
    \centering
    \includegraphics[scale=0.24]{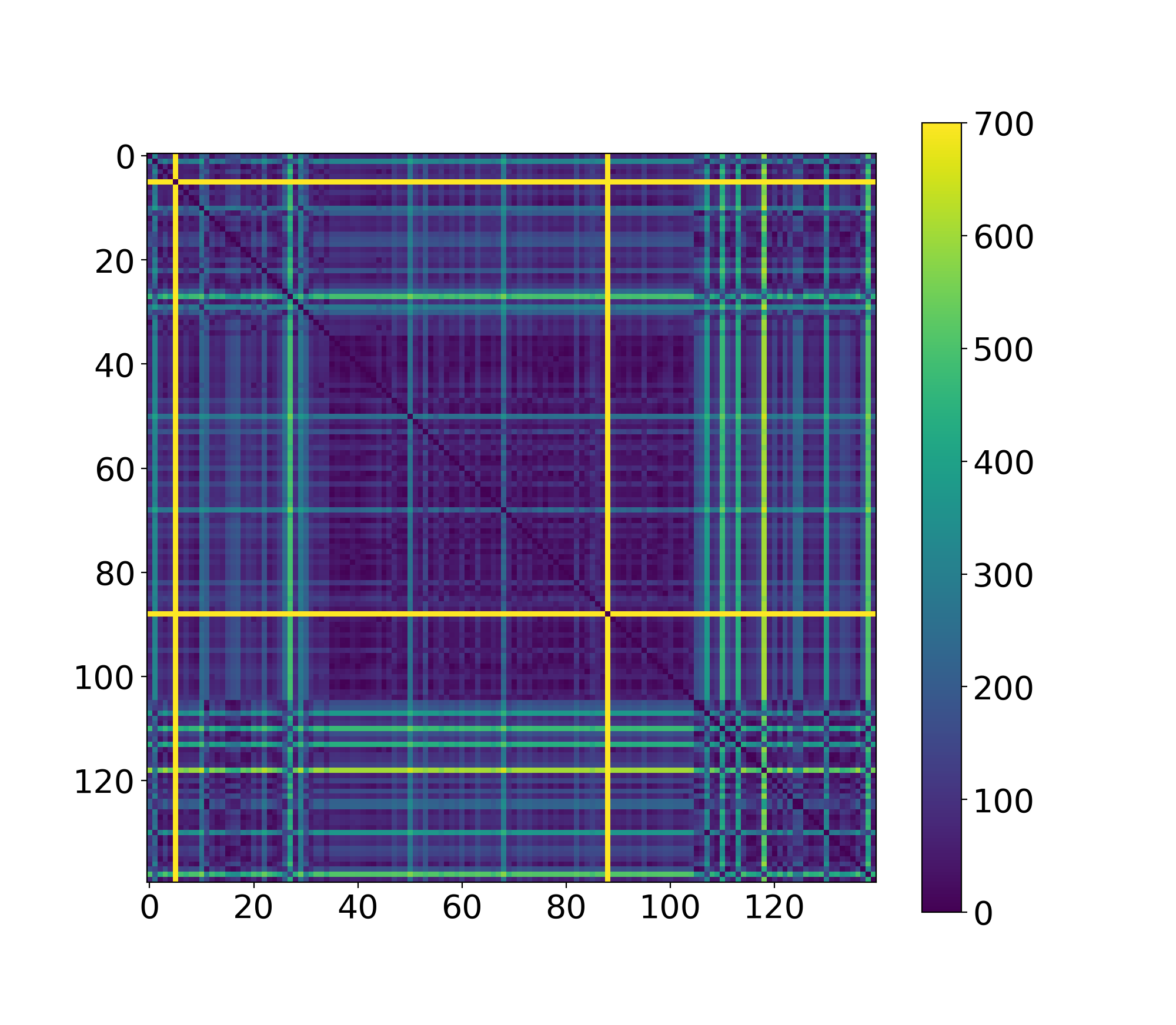}
    \includegraphics[scale=0.24]{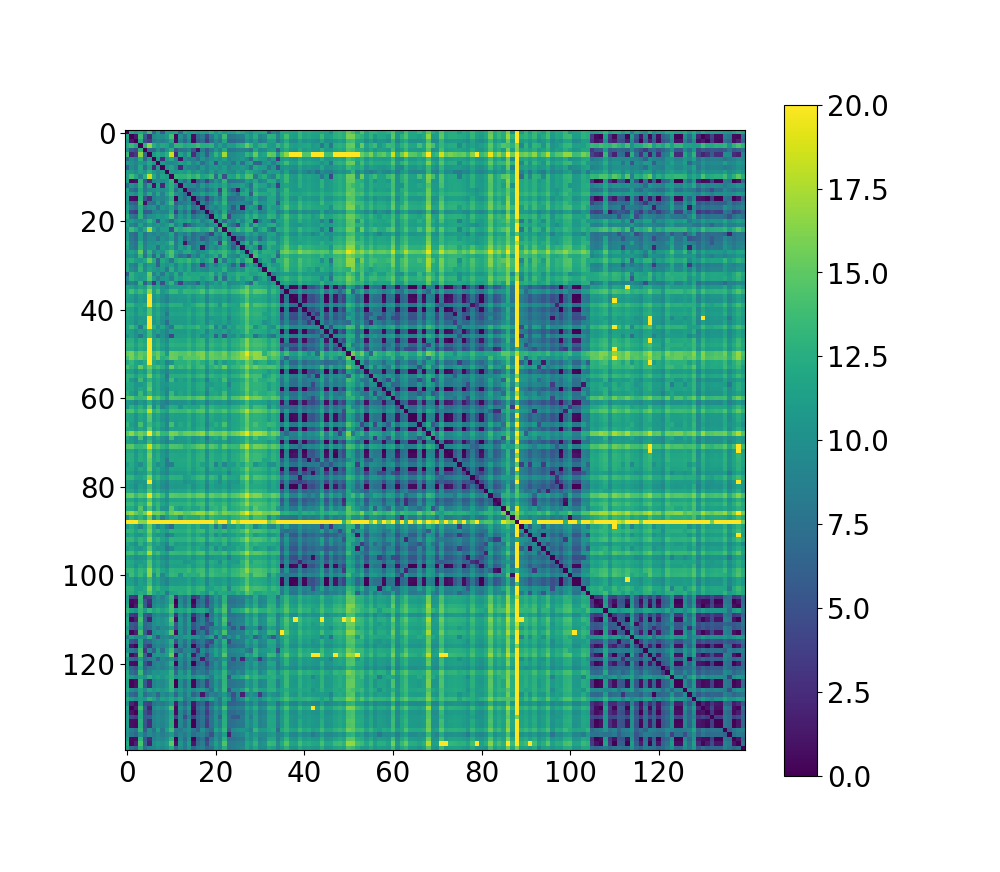}
    \caption{Distance matrices between the variational Dirichlet parameters of 140 documents from 4 classes of the 20NewsGroup dataset, for the Euclidean distance (left) and the Fisher-Rao geodesic distance (right). The indices are regrouped by classes, which are 'alt.atheism', 'comp.graphics', 'comp.os.ms-windows.misc', 'soc.religion.christian'.}
    \label{fig:text_distances}
\end{figure}

\begin{figure}
    \centering
    \includegraphics[scale=0.4]{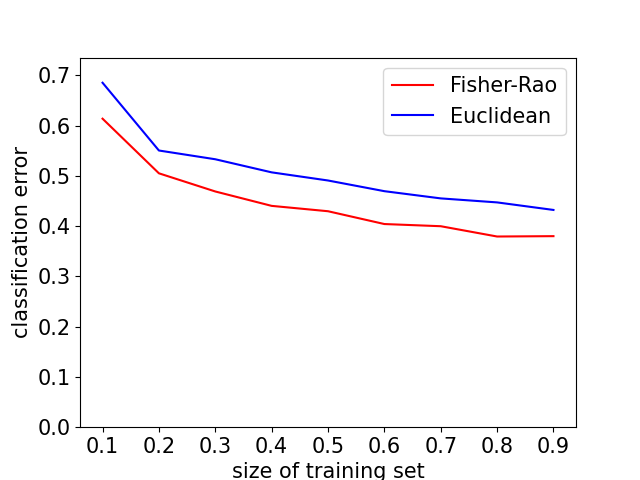}
    \caption{Classification error of K-nearest neighbors algorithm applied to 140 documents from 4 classes of the 20Newsgroups dataset, using the Euclidean distance and the Fisher-Rao geodesic distance, plotted with respect to the percentage of the data chosen for the training set.}
    \label{fig:text_knn}
\end{figure}

Here we also use the parameter space of Dirichlet distributions to represent documents. However, we use the Fisher-Rao metric instead of the Euclidean metric for comparison. We extract 140 documents from the 20Newsgroups dataset, a collection of news articles labeled according to their main topic. We select documents from 4 different classes: 'alt.atheism', 'comp.graphics', 'comp.os.ms-windows.misc', 'soc.religion.christian'. We then perform LDA on the obtained corpus, estimate the corresponding variational Dirichlet parameters on a space of $K=10$ topics, and use these to represent the documents in the $10$-dimensional parameter manifold of Dirichlet distributions. The pairwise distances between these parameters, regrouped by classes, for the Euclidean distance and the Fisher-Rao geodesic distance are shown in Figure~\ref{fig:text_distances}. While the 4 classes structure does not appear clearly, one can see 2 classes appear \textemdash one corresponding to religion and the other to computers \textemdash more distinctly with the Fisher-Rao metric than with the Euclidean metric. We use these distance matrices to perform $K$-nearest neighbors classification ($K=10$) after splitting the dataset into training and testing sets, and show the evolution of the classification error with respect to the percentage of data chosen for the training set in Figure~\ref{fig:text_knn}. We observe that the classification error is consistently lower for the Fisher-Rao metric compared to the Euclidean metric.

\section*{Conclusion}

In this paper, we presented a Python implementation of information geometry integrated in the software Geomstats. We showed that our module \codeobj{information\_geometry} contains the essential building blocks to perform statistics and machine learning on probability distributions data. As we have described the formulas and mathematical structures implemented in our module, we have also reviewed the main analytical results of the field and the main areas of applications. We also demonstrated a clear usecase of information geometry for text classification, where the geometry of the probability space helps improve the data analysis. We hope that our implementation will inspire researchers to use, and contribute to, information geometry with the Geomstats library.

\section*{Appendix}

\subsection{Proof of geodesic distance for geometric distributions}

A geometric distribution of parameter $p \in [0,1]$ has a p.m.f. : $$\forall k \geq 1,\, P(k|p) = f(k|p) = (1-p)^{k-1}p.$$
Then, for $0<p<1$, as $\frac{\partial^2 \log f}{\partial p^2} = \frac{1-k}{(1-p)^2} - \frac{1}{p^2}$, we have: 
$$I(p) = - \mathbb{E}_{p}\left[\frac{\partial^2 \log f(X)}{\partial p^2} \right] = \frac{1}{p^2} + \frac{\mathbbm{E}(X) -1}{(1-p)^2} =\frac{1}{p^2} + \frac{1}{p(1-p)} = \frac{1}{p^2(1-p)}$$
Then, with $ds$ the infinitesimal distance on the geometric manifold, we get:
$$ds^2 = \frac{1}{p^2(1-p)} dp^2.$$
Therefore the distance between $p_1$ and $p_2 \geq p_1$ writes: $$d(p_1, p_2) = \int_{p_1}^{p_2} \frac{1}{p} \frac{1}{\sqrt{1-p}}dp.$$With the change of variable $u = \sqrt{p}$, we eventually draw:
$$d(p_1, p_2) = 2 \int_{\sqrt{p_1}}^{\sqrt{p_2}}\frac{du}{u \sqrt{1 - u^2}} = 2 \left[\tanh^{-1}\left(\sqrt{1-u^2}\right)\right]_{\sqrt{p_1}}^{\sqrt{p_2}}.$$Finally: $$d(p_1, p_2) = 2\left(\tanh^{-1}\left(\sqrt{1-p_2}\right) - \tanh^{-1}\left(\sqrt{1-p_1}\right)\right).$$ 

\subsection{Proof of geodesic distance for on the Gamma manifold with fixed $\kappa$}

From the Fisher information matrix obtained in 3.4.1.2., we derive here: $$ds^2 = \frac{\kappa}{\gamma^2} d\gamma^2, $$and then for $\gamma_1 \leq \gamma_2$: $$d(\gamma_1, \gamma_2) = \sqrt{\kappa} \int_{\gamma_1}^{\gamma_2} \frac{d\gamma}{\gamma} = \sqrt{\kappa} \log\left(\frac{\gamma_2}{\gamma_1}\right).$$

\bibliographystyle{plain}
\bibliography{bibliography}

\end{document}